\pdfoutput=1
\documentclass[]{fairmeta}

\usepackage[round,authoryear]{natbib}
\usepackage{lineno} 
\usepackage{hyperref}
\usepackage[colorinlistoftodos]{todonotes}
\usepackage{colortbl}
\usepackage{nicematrix}
\usepackage{graphicx}
\usepackage{amssymb}
\usepackage{amsmath}
\usepackage{adjustbox}
\usepackage{soul}
\usepackage[inline]{enumitem}
\usepackage{booktabs}
\usepackage{color}
\usepackage{xcolor}
\usepackage{bbding}
\usepackage{listings}
\usepackage{multicol}
\usepackage{xspace}
\usepackage{lmodern}
\usepackage{tablefootnote}
\usepackage{times}
\usepackage{latexsym}

\usepackage{tabularx}
\usepackage[T1]{fontenc}
\usepackage[utf8]{inputenc}

\usepackage[utf8]{inputenc}

\usepackage{microtype}
\usepackage{inconsolata}

\usepackage{geometry}
\usepackage{makecell}
\usepackage{algorithm}
\usepackage{algorithmicx}
\usepackage{algpseudocode}
\usepackage{amsmath}
\usepackage{amsfonts}
\usepackage{booktabs}
\usepackage{subcaption}
\usepackage{adjustbox}
\usepackage{xcolor}
\usepackage{soul}

\newcommand{\approachname}{\textit{Soup Of Category Experts\space}}
\newcommand{\soce}{SoCE}

\makeatletter
\AtBeginDocument{

}
\makeatother
\title{Souper-Model: How Simple Arithmetic Unlocks State-of-the-Art LLM Performance}

\author[1,3\dagger]{Shalini Maiti}

\author[2\dagger]{Amar Budhiraja}
\author[2]{Bhavul Gauri}
\author[2]{Gaurav Chaurasia}
\author[2]{Anton Protopopov}
\author[2]{Alexis Audran-Reiss}
\author[2]{Michael Slater}
\author[2]{Despoina Magka}
\author[2]{Tatiana Shavrina}
\author[2,3\ast]{Roberta Raileanu}
\author[2]{Yoram Bachrach}

\affiliation[1]{Meta SuperIntelligence Labs}
\affiliation[2]{FAIR at Meta}
\affiliation[3]{University College London}

\contribution[\dagger]{Equal contribution}
\contribution[\ast]{Work done while working at Meta}
\abstract{

Large Language Models (LLMs) have displayed remarkable capabilities across diverse domains, but their training remains resource- and time-intensive, requiring massive computational resources and careful orchestration of training procedures. Model souping—the practice of averaging weights from multiple models of the same architecture—has emerged as a promising pre- and post-training technique that can enhance performance without expensive retraining. We observe that previous souping approaches can lead to collapse in precision-sensitive LLM capabilities. 

In this paper, we introduce \approachname(\soce), a principled approach for model souping to overcome this shortcoming. The proposed method utilizes benchmark composition to identify optimal model candidates and applies non-uniform weighted averaging to maximize performance. Contrary to previous approaches, our method leverages the observation that different clusters (or categories) of points within a benchmark often exhibit low inter-correlations in model performance. \soce{} identifies "expert" models for each weakly-correlated category cluster and combines them using optimized weighted averaging rather than uniform weights. We demonstrate that \soce{} improves performance and robustness across multiple domains and achieves state-of-the-art results on the Berkeley Function Calling Leaderboard.

}

\date{\today}
\correspondence{Shalini Maiti at \email{shalini.maiti@gmail.com}, Amar Budhiraja at \email{amar.budhiraja1@gmail.com}}

\metadata[Code]{\url{https://github.com/facebookresearch/llm_souping}}

\crefformat{section}{\S#2#1#3}
\crefformat{subsection}{\S#2#1#3}
\crefformat{subsubsection}{\S#2#1#3}

\definecolor{pastelgreen}{RGB}{102,204,102}
\definecolor{pastelred}{RGB}{255,77,77}

\begin{document}
\maketitle
\section{Introduction}

Large Language Models (LLMs) have emerged as transformative technologies, demonstrating remarkable capabilities across diverse domains from natural language understanding to code generation and tool use~\citep{brown2020language,comanici2025gemini,dubey2024llama,roziere2023code,schick2023toolformer,llama,touvron2023llama2,achiam2023gpt,team2023gemini}. The training paradigms for these foundational models typically involve massive-scale pretraining on diverse corpora followed by supervised fine-tuning and reinforcement learning from human feedback~\citep{ouyang2022training,bai2022training}. However, current training procedures remain extremely resource and time-intensive, often requiring large amounts of compute power and careful orchestration of training data mixtures to achieve desired competencies in the models~\citep{hoffmann2022training,touvron2023llama2,zhang2022opt}.

Recent advances in model souping—averaging weights from multiple models of the same architecture—have shown promising results as a post-training technique to enhance performance without requiring expensive retraining. Prior work has demonstrated that uniform averaging of model weights can lead to improved performance across various tasks~\cite{wortsman2022model} and mitigate issues like catastrophic forgetting~\cite{kleiman2025soup}. These findings suggest that model souping offers a computationally efficient alternative to traditional approaches that rely on modifying training data recipes or retraining procedures. Recently, the concept of "model souping" has evolved into a broader model merging paradigm that combines the weights of multiple task-specific derivative models into a single, highly capable network without the need for additional training. TIES-Merging ~\citep{ties_yadav2023tiesmerging}, an example of such an approach, isolates the fine-tuning updates and resets up to 80\% of the smallest parameter changes to zero, retaining only the most significant delta weights. DARE ~\citep{yu2024language} proposes randomly dropping a vast majority of these fine-tuned delta weights and rescaling the remaining parameters to maintain their overall mathematical expectation. While these approaches have been shown to push performance, our experiments show that these approaches can collapse several LLM capabilities which require exact instruction adherence, such as tool-use, long-context, and reasoning (Tables \ref{ties_dare_other_benchmarks_8b}, \ref{tab:ties_dare_other_benchmarks_bfcl_per_cat_comp} and \ref{ties_dare_other_benchmarks_70b}). To address these concerns, we introduce an improved dense model averaging method that combines benchmark-guided candidate selection with non-uniform weighting to form a model soup leading to state-of-the-art on the Berkeley Function Calling Leaderboard (BFCL)~\citep{BFCL}. The contributions of this paper are as follows:
\begin{enumerate}
    \item \textbf{Benchmark Aware Souping}: We introduce \approachname (\soce), a novel model souping technique that leverages benchmark composition through category-aware expert selection mechanism. Unlike previous approaches, \soce{} doesn't lead to collapse of precision-sensitive capabilities like tool-use.
    \item \textbf{State-of-the-Art Performance:} We demonstrate the efficiency of the proposed method across diverse domains, including state-of-the-art results for the BFCL~\citep{BFCL}. Our approach outperforms existing baselines, validating the effectiveness of category-specific model souping.
    \item \textbf{Improved Model Consistency:} We perform a large-scale empirical analysis to show that model souping enhances performance consistency across benchmark categories. Souped models exhibit significantly higher Pearson correlations between category performances within a benchmark across model populations compared to their unsouped counterparts, indicating improved robustness and coherence across diverse task types.
\end{enumerate}
We also discuss the connection of model souping with game theory and Shapley values; revisit the assumptions of existing model merging approaches in the context of specialized models; present qualitative analysis for BFCL experiments; understand impact on other benchmarks when souping for a given benchmark and present additional results on FLORES ~\citep{flores101_goyal2021} multi-lingual benchmark as part of appendix to this paper. 
\section{Related Work}

\begin{table*}[htbp]
\centering
\resizebox{\textwidth}{!}{%
\begin{tabular}{llllll}
\toprule
\textbf{Property} & \makecell[l]{\textbf{SoCE}\\\textbf{(Proposed Approach)}} & \makecell[l]{\textbf{TIES}\\\cite{ties_yadav2023tiesmerging}} & \makecell[l]{\textbf{DARE}\\\cite{yu2024language}} & \makecell[l]{\textbf{Model Soups}\\\cite{wortsman2022model}} & \makecell[l]{\textbf{Model Stock}\\\cite{jang2025modelstock}} \\
\midrule
\textbf{Operates on} 
    & Raw LLM weights 
    & Task vectors 
    & \makecell[l]{Task vectors\\(preprocessor)} 
    & Raw weights 
    & \makecell[l]{Raw weights\\(layer-wise)} \\
\addlinespace
\textbf{Candidate selection} 
    & \makecell[l]{Yes\\(correlation-guided)} 
    & No 
    & No 
    & \makecell[l]{No\\(all or greedy on val)} 
    & \makecell[l]{No\\(fixed: 2 models)} \\
\addlinespace
\textbf{Weight optimisation} 
    & \makecell[l]{Non-uniform\\grid search} 
    & \makecell[l]{Majority vote\\on signs} 
    & \makecell[l]{Rescale by\\$1/(1-p)$} 
    & \makecell[l]{Uniform or\\greedy} 
    & \makecell[l]{Closed-form\\(angle $\theta$)} \\
\addlinespace
\textbf{Data needed at merge} 
    & \makecell[l]{Benchmark\\eval scores} 
    & None 
    & None 
    & \makecell[l]{Held-out val set\\(greedy)} 
    & None \\
\addlinespace
\textbf{Modifies parameters} 
    & \makecell[l]{Yes\\(weighted avg)} 
    & \makecell[l]{Yes\\(trim + sign elect)} 
    & \makecell[l]{Yes\\(random zero + rescale)} 
    & \makecell[l]{Yes\\(averaged weights)} 
    & \makecell[l]{Yes\\(weighted avg, layer-wise)} \\
\addlinespace
\textbf{Training required} 
    & No 
    & No 
    & No 
    & No 
    & \makecell[l]{No\\(periodic merging optional)} \\
\addlinespace
\textbf{Interference handling} 
    & \makecell[l]{Complementary\\expert selection} 
    & \makecell[l]{Trim + sign\\resolution} 
    & \makecell[l]{Sparsification\\of deltas} 
    & \makecell[l]{None\\(relies on shared basin)} 
    & \makecell[l]{Geometric anchoring\\to pretrained} \\
\addlinespace
\textbf{Formal guarantee} 
    & No 
    & No 
    & \makecell[l]{Yes\\(in expectation)} 
    & Partial 
    & \makecell[l]{Yes\\(given angular consistency)} \\
\addlinespace
\textbf{Scalability with $N$} 
    & \makecell[l]{Scales\\(selects subset)} 
    & \makecell[l]{Better than TA\\as $N$ grows} 
    & \makecell[l]{Plug-in,\\scales with host} 
    & \makecell[l]{Scales\\(greedy filters)} 
    & \makecell[l]{Designed for $N=2$;\\extends to $N$} \\
\addlinespace
\makecell[l]{\textbf{Risk to critical neurons}\\\cite{achillesheel_qin2026}} 
    & \makecell[l]{Low\\(conservative avg)} 
    & \makecell[l]{Higher\\(trim discards)} 
    & \makecell[l]{Higher\\(random zeroing)} 
    & \makecell[l]{Low\\(conservative avg)} 
    & \makecell[l]{Low\\(pretrained anchor preserves)} \\
\addlinespace
\textbf{Models needed} 
    & Subset of candidate pool 
    & $\geq 2$ 
    & $\geq 2$ (preprocessor) 
    & $\geq 2$ (typically many) 
    & Exactly 2 \\
\addlinespace
\textbf{Domain} 
    & (V)LLM (Complex benchmarks) 
    & Vision + NLP 
    & NLP (LLM-focused) 
    & Vision + NLP 
    & Vision (CLIP) \\
\addlinespace
\textbf{Pretrained model used} 
    & No 
    & As origin (implicit) 
    & As origin (implicit) 
    & No 
    & Yes (explicit geometric anchor) \\
\addlinespace
\textbf{Category structure} 
    & \textbf{by within-benchmark} 
    & diverse tasks
    & diverse tasks 
    & diverse tasks         
    & diverse tasks \\
\addlinespace
\bottomrule
\end{tabular}%
}
\caption{Comparison of \soce{} and other Model Merging methods. Refer to Table ~\ref{tab:rw-comparison-extended} for more methods.}
\label{tab:rw-comparison}
\end{table*} 

We outline the current souping/merging methodologies that have emerged to improve LLM efficiency and generalization to contextualize \soce{}'s novelty and usefulness within this space.

\textbf{Souping:} Starting with \citet{Utans1996WeightAF}, model Souping has been explored in the machine learning literature in multiple contexts~\cite{jang2025modelstock,jaiswal2023instant,kleiman2025soup,wortsman2022model,yu2024language,zimmer2023sparse}. Weight averaging found theoretical foundation with ~\citep{izmailov2018averaging}, that showed that averaging checkpoints along a stochastic gradient descent trajectory finds flatter, better-generalizing minima; \citet{wortsman2022model} extended this idea across independently fine-tuned models, showing how souping on model variants across different hyperparameter configurations of the same base model leads to improvements over individual candidates. They proposed 3 strategies: Uniform (where all models are weighted equally), Greedy (add 1 model at a time in decreasing order of performance) and Learned-Souping (soup weights are learned via gradient descent), with Greedy Souping yielding the most promising results. ~\citet{wortsman2022wise} further showed that interpolating between zero-shot and fine-tuned weights improves robustness under distribution shift, establishing weight-space interpolation as a practical regularization. ~\citet{jang2025modelstock} improve upon the work by \citet{wortsman2022model} using a geometric insight - proximity to pre-trained models - to reduce the number of models to two to achieve better results than its predecessor. \soce{} takes it further by utilizing the insights from the correlation of metrics within a benchmark and co-operative game theory to formalize the selection of the best candidates and their souping proportion. The viability of weight averaging rests on the theoretical observation that models fine-tuned from the same pretrained weights remain in the same loss basin~\citep{neyshabur2020what} and exhibit linear mode connectivity~\citep{frankle2020linear} - preconditions naturally satisfied by all candidate models in \soce{}, which share a common pretrained initialization. \citet{rame2022diwa} further provided a formal bias-variance-covariance decomposition showing that weight averaging succeeds when model diversity is high, reducing covariance between model errors - the condition \soce{} engineers by selecting anti-correlated category experts. In a related study~\cite{yu2024language}, the authors take several homologous (same architecture and parameters) models and perform deactivation and reactivation of neurons and merge the newly derived models to achieve better models compared to the input candidates. Model souping has also been used in continual learning to address catastrophic forgetting during finetuning of LLMs~\cite{kleiman2025soup}, and in RLHF~\citep{rame2024warm} where weight averaging of reward models has been shown to improve alignment. In LLMs~\cite{dubey2024llama,cohere2025commandaenterprisereadylarge}, it has been used as part of the fine-tuning pipeline. A principal distinction between all prior souping approaches and the method introduced here lies in their respective objectives: previous work has primarily leveraged souping to achieve inter-benchmark improvements - combining models specialised in coding and function calling~\citep{cohere2025commandaenterprisereadylarge} or across 16 specialised domains~\citep{ablin2025soup} - whereas \soce{} systematically applies souping to enhance performance on a single benchmark by exploiting its internal compositional structure through Pearson correlation analysis of subcategory scores, encapsulating capability complementarity directly into the candidate selection process itself.

\textbf{Automatic model merging:} Several works~\cite{matena2022mergingmodelsfisherweightedaveraging,stoica2024zipitmergingmodelsdifferent,ties_yadav2023tiesmerging,ablin2025soup} explore automatic model merging techniques. \citet{task_arithmetic_ilharco2023editing} introduced task arithmetic, defining task vectors as the difference between fine-tuned and pretrained weights and showing these vectors can be added, negated, or combined to edit model behaviour - however, this suffers from interference when models push shared parameters in conflicting directions. \citet{ties_yadav2023tiesmerging} proposed TIES-Merging to address this - redundant low-magnitude parameters and sign conflicts - and resolving them through a trim-elect-merge procedure, while \citet{yu2024language} introduced DARE, demonstrating that 90-99\% of delta parameters can be randomly dropped and rescaled without degrading performance, substantially reducing interference before merging. ~\citet{matena2022mergingmodelsfisherweightedaveraging} proposed Fisher-weighted averaging using the Fisher information matrix to assign per-parameter importance weights, and \citet{yang2024adamergingadaptivemodelmerging} introduced AdaMerging, which discovers optimal layer-wise merging coefficients via entropy minimization without requiring labeled data, with an application to classification and ViT models. \citet{jin2023dataless_regmean} developed RegMean, a closed-form merging solution via inner product regression. \citet{Akiba_2025} apply evolutionary algorithms to model merging to find high-performing multi-modal LLMs. We provide a tabular summary of the properties of \soce{} versus other merging methods in Tables ~\ref{tab:rw-comparison}, ~\ref{tab:rw-comparison-extended}. Existing merging methods share certain characteristics that make them difficult to apply to current LLMs: requiring either privileged access to training data or activation statistics~\cite{jin2023dataless_regmean}, impose architectural constraints~\cite{slerp_jang2024}, degrade as the number of merged models grows~\cite{task_arithmetic_ilharco2023editing}, or risk destroying catastrophically critical neurons identified by~\citet{achillesheel_qin2026} through aggressive parameter trimming/random zeroing~\citep{ties_yadav2023tiesmerging,yu2024language}. Even \citet{jang2025modelstock} and \citet{wortsman2022model} select candidates based purely on overall validation accuracy rather than structured capability coverage - whereas \soce{} uniquely positions itself to avoid such failure modes by operating entirely in benchmark performance space, requiring only publicly available evaluation scores, and preserving all weights through conservative weighted averaging that minimizes critical neuron disruption.

\FloatBarrier

\section{Methodology: \approachname}
\begin{figure}[H]
  \centering
  \includegraphics[width=0.85\linewidth]{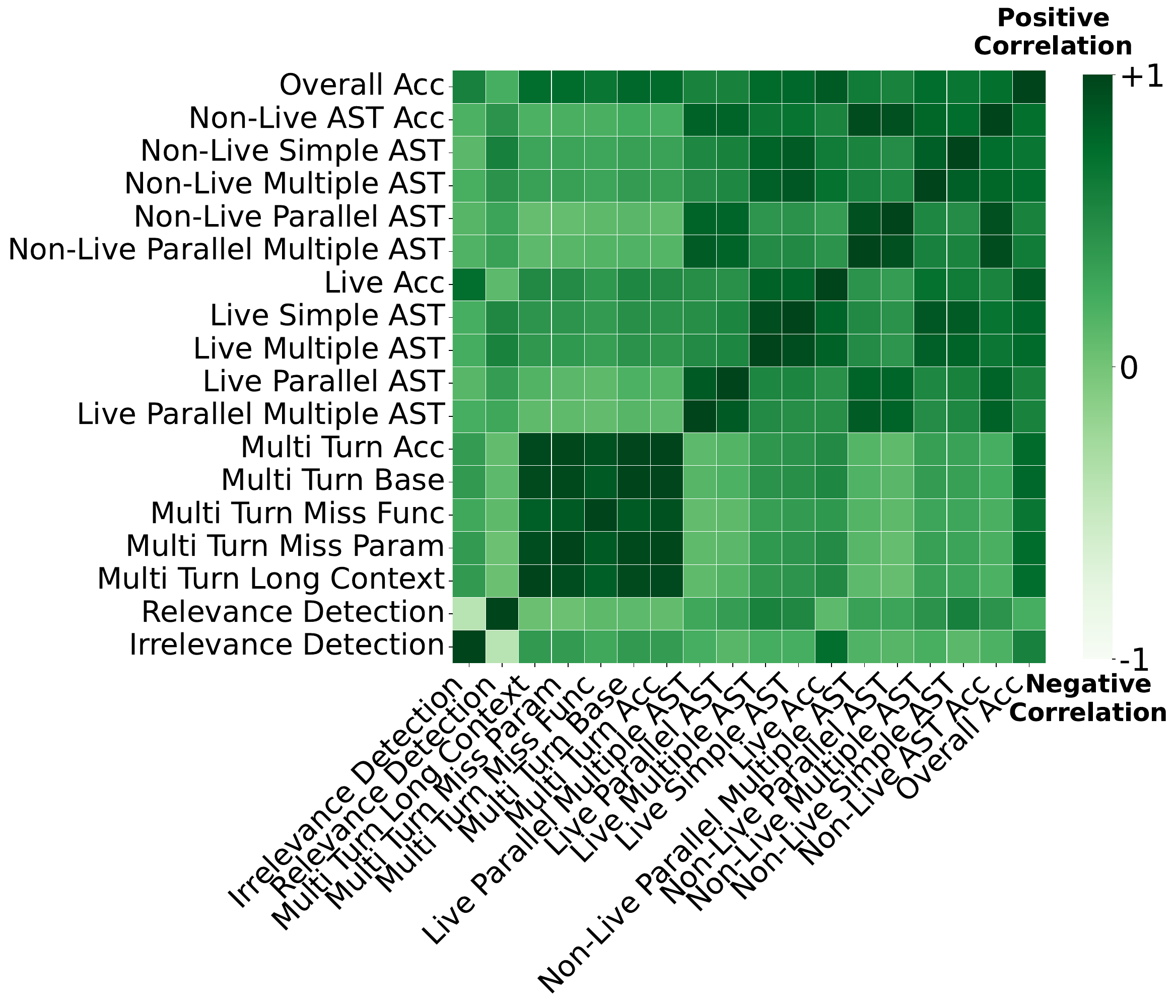}
  \caption{Pearson Correlation of models' performance on BFCLv3.}
  \label{fig:bfcl_leaderboard_pearson_corr}
\end{figure}

In this section, we present our method for  model selection and weighting the selected models  with the benchmark composition in mind.

The fundamental insight underlying our approach is that benchmark performance across categories exhibits heterogeneous correlation patterns. Different models demonstrate varying expertise across benchmark categories, with some categories being strongly correlated while others remain weakly correlated or even negatively correlated in terms of cross-model performance. To illustrate this phenomenon, we analyze the Berkeley Function Calling Leaderboard (BFCLv3) \cite{BFCL}, which comprises multiple categories including multi-turn function calling, irrelevance detection, and function calling across different programming languages (Java, Javascript, etc.). Figure \ref{fig:bfcl_leaderboard_pearson_corr} presents a correlation heatmap showing Pearson correlation coefficients between category performances across all models on the leaderboard, where darker regions indicate higher correlations. The heatmap reveals strong positive correlations (dark green regions) between related categories and weak to negative correlations (light green regions) between unrelated categories. For instance, multi-turn categories exhibit high inter-correlation (0.96-0.98), indicating that models proficient in one multi-turn task typically excel across all multi-turn scenarios. Conversely, weak correlation (0.07) exists between Multi-turn-base (where the model is evaluated on Multi-turn function calling aspect) and Live Accuracy (where the model is evaluated on real-world user collected function calling prompts) categories, suggesting these represent distinct competency domains.

\begin{algorithm}
\small
\caption{\approachname (\soce)}
\label{alg:sose}
\begin{algorithmic}[1]

\Require Benchmark dataset $\mathcal{D}$ with categories $\{C_1, C_2, \ldots, C_k\}$
\Require Set of candidate models $\mathcal{M} = \{M_1, M_2, \ldots, M_n\}$

\Require Correlation threshold $\tau$ for identifying low-correlation categories
\Ensure Souped model $M_{\text{soup}}$

\State \textbf{Step 1: Analysis}

\For{each pair of categories $(C_i, C_j)$ where $i \neq j$}
    \State Compute Pearson correlation $\rho_{i,j}$ between $(P^i_1,...,P^i_n)$ and $(P^j_1,...,P^j_n)$ across all models  and category performances
\EndFor

\State $L = \left\{ C_k \;\middle|\; \exists\, l \text{ such that } |\rho_{k,l}| < \tau \text{ or } |\rho_{l,k}| < \tau \right\}$

\State \textbf{Step 2: Expert Model Selection}

\For{each category $C_i \in L$}
    \For{each model $M_j \in \mathcal{M}$}
        
        \State $P_j^i = \text{Performance}(M_j, C_i)$
    \EndFor
    \State Select expert model $M^*_i = \arg\max_{M_j \in \mathcal{M}} P_j^i$
\EndFor

\State \textbf{Step 3: Weight Optimization}
\State Apply uniform grid search to find optimal weights, such that for $\mathbf{w} = \{w_1, \ldots  w_l\}$ it is $\sum_{i=1}^l w_i = 1$, $l =|L|$ and
$\mathbf{w}^* = \arg\max_{\mathbf{w}}$
\Statex \hspace{\algorithmicindent}  $\frac{1}{k} \sum_{i=1}^k \text{Performance}\left(\sum_{j=1}^l w_j \cdot M^*_j, C_i\right)$
\State \textbf{Step 4: Souped Model}
\State Create souped model: $M_{\text{soup}} = \sum_{i=1}^l w^*_i \cdot M^*_i$
\State \Return $M_{\text{soup}}$
\end{algorithmic}
\end{algorithm}

Our proposed method, \approachname (\soce), exploits these correlation patterns to strategically select and weight models for souping. The core principle is to identify expert models for each weakly-correlated category cluster and aggregate them using optimized weighted averaging to combine complementary expertise.  Algorithm \ref{alg:sose} formalizes this procedure, which comprises four key steps: (1) correlation analysis to identify weakly-correlated category pairs, (2) expert model selection for each category based on performance rankings, (3) weight optimization to maximize aggregate performance, and (4) weighted model souping to produce the final combined model. For weight optimization, we perform a search over a uniform set of weights. We iterate over all combinations in the weight space with a maximum weight value of 0.9, lowest weight value of 0.1 and step size of 0.1, where each weight corresponds to a single model . We also add a special case of equal weighing of the candidates to compare uniform souping.  We discuss the implications of the correlation threshold ($\tau$) on the proposed approach in Section \ref{tau_appendix}.

\section{Experiments}
\begin{table*}[h!]
    \centering
    \begin{subtable}[t]{0.48\linewidth}
        \centering
        \begin{tabularx}{\linewidth}{|X|l|}
            \hline
            \textbf{Model} & \makecell[l]{\textbf{BFCL} \\ \textbf{Accuracy}}  \\
            \hline
            \makecell[l]{xLAM-2-70b\\~\citep{prabhakar2025apigen}} & \underline{78.56\%}  \\
            \makecell[l]{CoALM-70B\\~\citep{acikgoz2025can}} & 54.49\%  \\
            \makecell[l]{watt-tool-70B\\~\citep{wattai_70b}} &  73.57\% \\
            \makecell[l]{functionary-medium-70B\\~\citep{fc_70b}} &  62.32\% \\
            \hline 
            \makecell[l]{Uniform Souping\\~\citep{wortsman2022model}} &  68.33\% \\
            \hline
            \makecell[l]{Uniform Souping with\\ \soce\space Model Selection} &  78.40\% \\
            \hline
            \makecell[l]{\soce\\(Proposed Method)}  & \textbf{80.68\%}  \\
            \hline
        \end{tabularx}
        \caption{70 billion parameters models}
        \label{table:bfcl70bresult}
    \end{subtable}
    \hfill
    \begin{subtable}[t]{0.48\linewidth}
        \centering
        \begin{tabularx}{\linewidth}{|X|l|}
            \hline
            \multicolumn{1}{|c|}{\textbf{Model}} & \textbf{BFCL Accuracy}  \\
            \hline
            \makecell[l]{xLAM-2-8b\\~\citep{prabhakar2025apigen}} &  \underline{72.37\%}  \\
            \makecell[l]{ToolACE-2-8B\\~\citep{liu2024toolace}} & 68.73\% \\
            \makecell[l]{watt-tool-8B \\~\citep{wattai_8b}} & 67.79\%   \\
            \makecell[l]{BitAgent-8B\\~\citep{bitagent_8b}} & 67.49\%\\
            \makecell[l]{CoALM-8B\\~\citep{acikgoz2025can}} & 54.52\% \\ 
            \hline 
            \makecell[l]{Uniform Souping\\~\citep{wortsman2022model}} & 69.80\% \\
            \hline
            \makecell[l]{Uniform Souping with\\ \soce\space Model Selection} & 74.01\% \\
            \hline
            \makecell[l]{\soce\\(Proposed Method)}  & \textbf{76.50\%}  \\
            \hline
        \end{tabularx}
        \caption{8 billion parameter models}
        \label{table:bfcl8bresult}
    \end{subtable}
    \caption{BFCL Performance of 8 billion and 70 billion parameter models. The first 4 entries are the ingredient models and the last 3 are Souped Models. Uniform Souping refers to the baseline where all models are combined together with the same weight, Uniform Souping with model selection only combines the selected models using the anti-correlation criterion and \soce\space is the proposed method that augments Uniform Souping with model selection with weights optimization.}
    \label{table:bfcl-combined}
\end{table*}





\subsection{Benchmarks and Baselines}
We conduct comprehensive experiments to evaluate the effectiveness of our proposed \approachname (\soce) methodology across three diverse benchmarks that span different LLM capabilities:
\begin{enumerate}
    \item \textbf{Berkeley Function Calling Leaderboard (BFCL)} ~\cite{BFCL}: evaluates tool calling and function invocation capabilities of LLMs across multiple categories including multi-turn interactions, irrelevance detection, and cross-language function calling. We use BFCL-v3 commit id 1bb65c9 for evaluations as stated in the public BFCL-v3 leaderboard.
    \item \textbf{Multilingual Grade School Math Benchmark (MGSM)} ~\citep{mgsm_shi2022}: assesses mathematical reasoning abilities across multiple languages, testing both computational skills and cross-lingual generalization.
    \item \textbf{$\infty$Bench}~\citep{infinity-bench}: evaluates long-context processing capabilities, testing models' ability to maintain coherence and extract information from extended sequences. We use a subset containing 3 categories.
\end{enumerate}

For all benchmarks, we compare individual models with the following model soups: 
\begin{enumerate}
    \item \textbf{Uniform Souping (All Candidate Models)} ~\cite{wortsman2022model}: Baseline approach uniformly averaging all candidate models.
    \item \textbf{Uniform Soup with \soce{} Model Selection}: Uniform weighting applied to our selected candidates based on anti-correlated categories.
    \item \textbf{\soce{} (Weighted Souping with Model Selection)}: Complete proposed method with candidate selection and optimized weighting.
\end{enumerate}
This experimental design enables targeted ablation studies: comparing (1) vs (2) isolates our candidate selection benefits, (2) vs (3) quantifies weight optimization gains, and (1) vs (3) demonstrates overall \soce{} performance improvements. We also compare current state-of-the-art individual models to quantify improvements of souping over these models. We also evaluate \soce{} against TIES ~\citep{ties_yadav2023tiesmerging}, DARE (DARE-Linear) ~\citep{yu2024language} and DARE-TIES ~\citep{yu2024language} on BFCL. 

\begin{table}[t]
\centering
\label{tab:8b_results}
\resizebox{0.5\columnwidth}{!}{%
\begin{tabular}{ll ccccc}
\toprule
\textbf{Method} & \textbf{$d$} & \textbf{TQA} & \textbf{IFEval} & \makecell{\textbf{MMLU-} \\ \textbf{Pro}} & \makecell{\textbf{Long-} \\ \textbf{Bench}}& \textbf{BFCL} \\
\midrule
ToolACE-8B & — & 53.3 & \textbf{67.8} & 34.1 & \textbf{30.9} & 68.73 \\
xLAM-8B & — & 54.2 & 28.5 & 33.1 & 24.4 & 72.37 \\
watt-8B & — & 54.1 & 56.0 & 34.2 & 22.7 & 67.79 \\
\midrule
TIES & 0.2 & \textbf{55.0} & 48.8 & 8.5 & 5.9 & 43.8 \\
DARE-Linear & 0.3 & 54.8 & 49.7 & 7.7 & 6.3 & 44.4 \\
DARE-TIES & 0.5 & 54.5 & 47.3 & 7.6 & 5.7 & 43.4 \\
\midrule
SoCE & — & 54.3 & 64.0 & \textbf{38.9} & 24.7 & \textbf{76.5} \\
\bottomrule
\end{tabular}}
\caption{Performance of Merging Methods on 8B models: Comparison of \soce{} against TIES and DARE across reasoning, instruction following, and function calling. Bold values indicate best-in-category. All metrics are reported in percentages. TIES and DARE exhibit significant regressions in MMLU-Pro and LongBench, whereas \soce{} maintains or improves performance relative to SOTA specialized models.}
\label{ties_dare_other_benchmarks_8b}
\end{table}
\subsection{Numerical Takeaways}
\label{numerical_results}
We compare souping for two model lines on BFCL: 70 billion and 8 billion parameter dense models. We identified 4 70B model candidates from the official leaderboard and applied \soce{} to them. \soce{} achieved 80.68\% accuracy, establishing a new state-of-the-art with a 2.84\% improvement over the previous best-performing individual model, xLAM-2-70b-fc-r~\citep{prabhakar2025apigen} (78.45\%). The optimal configuration utilized xLAM-2-70b-fc-r, CoALM-70B~\citep{acikgoz2025can}, and watt-tool-70B~\cite{wattai_70b} with 0.5, 0.2 and 0.3 weights, respectively. For 8B models, \soce{} achieved 76.50\% accuracy, surpassing the previous state-of-the-art within the 8B model size, xLAM-2-8b-fc-r~\citep{prabhakar2025apigen} by 5.7\% relative, with optimal weights of 0.7, 0.2, and 0.1 for xLAM-2-8b-fc-r, ToolACE-2-8B~\citep{liu2024toolace}, and watt-tool-8B~\citep{wattai_8b} respectively. For ablation studies on \soce{}'s model selection, we compare Uniform Souping (all candidate models) with Uniform Souping with \soce{} model selection, and observe that model selection for both 70B and 8B boosts performance. We further compare Uniform Souping with \soce{} model selection with \soce{} to understand the impact of weight optimization on the model soup, and observe that it leads to a relative improvement of 2.91\% for 70B model and 3.36\% for 8B models.

We evaluated TIES, DARE-Linear, and DARE-TIES against \soce{} to quantify the impact of weight-merging on specialized capabilities. As shown in Table \ref{table:bfcl-combined}, TIES and DARE approaches suffer from catastrophic performance degradation for BFCL achieving 0\% success on the Multi-Turn category of BFCL. This hints at models' long context capability collapse post TIES/DARE merging.
To validate this further, we benchmark these approaches on four other benchmarks using LM Evaluation Harness~\citep{eval_harness}: Truthful Q/A ~\cite{lin2022truthfulqa} (0-shot MC2 Accuracy), LongBench~\citep{bai2024longbench} (0-shot average subtask score), MMLU-Pro~\citep{wang2024mmlupro} (5-shot Exact Match), and IFEval~\citep{zhou2023instruction} (0-shot Strict Prompt Accuracy). We present these results for 8b models in Table \ref{ties_dare_other_benchmarks_8b}. While TruthfulQA remains stable and even gets a minor performance boost, MMLU-Pro and LongBench scores collapse to $<10\%$, indicating a fundamental failure in maintaining the parameter calibration required for complex reasoning and long-context processing. In contrast, \soce{} achieves a superior 76.5\% on BFCL and 38.9\% on MMLU-Pro, outperforming both the merging baselines and the specialized models. While \soce{} is not best at LongBench or TruthFul Q/A compared other models, it can be noted that its better than the best BFCL individual ingredient model (xLAM-8b) and comparable to other models.  These results demonstrate that \soce{} effectively synthesizes ingredient model capabilities without the reciprocal inhibition or erosion characteristic of previous merging techniques. We observe similar results for 70 billion-sized model for TIES and DARE approaches across these benchmarks and present the same in Appendix \ref{70b_dare_ties_section}.

We present the results on the Multilingual Grade School Math (MGSM) Benchmark ~\citep{mgsm_shi2022} in Table \ref{mgsm-70b-sose}. Similar to BFCL, we observed that \soce{} performs better than the candidate models and uniform souping.  We consider 4 open source models with 6.74 billion parameters for model souping: MetaMathOctopus-7B ~\citep{chen2023breaking}, 
MetaMathOctopus-MAPO-DPO-7B~\citep{she2024mapo}, MathOctopus-MAPO-DPO-7B~\citep{she2024mapo}, and 
Mathoctopus-Parallel-7B~\citep{she2024mapo}. We present the results for checkpoints synthesized using uniform souping~\citep{wortsman2022model}, uniform souping with \soce \space candidate selection, and \soce{} in Table \ref{mgsm-70b-sose}. Uniform souping, considering all candidates, leads to performance regression compared to the best candidate models. 

We also perform similar ablations to $\infty$Bench~\cite{infinity-bench} by training 5 Llama 3 architecture checkpoints~\citep{dubey2024llama} with 70 billion parameters with variations of the same data mix to understand if souping is useful beyond tool calling and math for LLMs. We present these results in  Tables \ref{table:infitity_bench_ablation} and \ref{tab:infbench-sub}. We can see that even though the candidate models have similar performance having been trained on variants of a single data mix, model souping is still helpful in improving performance. Uniform souping does not regress performance, and uniform souping with \soce{} model selection improves performance by 1.49\% relatively. Further, \soce{} led to a 2.04\% relative (0.56\% absolute) lift in performance compared to the best model candidate, also demonstrating that both weight tuning and candidate selection are important.

\begin{table}[t]
\centering
\begin{minipage}{0.48\linewidth}
\small
\centering
\begin{tabularx}{\linewidth}{|X|l|}
\hline
\multicolumn{1}{|c|}{\textbf{Model}} & \makecell{\textbf{MGSM} \\ \textbf{Accuracy}} \\
\hline
\makecell{MetaMathOctopus-7B \\~\citep{chen2023breaking}} & 41.9\% \\
\makecell{MetaMathOctopus-MAPO-\\DPO-7B~\citep{she2024mapo}} & \underline{50.9\%} \\
\makecell{MathOctopus-MAPO-DPO-7B\\~\citep{she2024mapo}} & 39.0\% \\
\makecell{Mathoctopus-Parallel-7B\\~\citep{she2024mapo}} & 35.5\% \\
\hline
\makecell{Uniform Soup} & 47\% \\
\hline
\makecell{Uniform Soup w/ \soce{} \\Model Selection} & 47.8\% \\
\hline
\makecell{\soce{} (Proposed Method)} & \textbf{51.7\%} \\
\hline
\end{tabularx}
\vspace{-0.25cm}\captionof{table}{Results for MGSM Benchmark: First 4 rows are ingredient models and the last 3 rows are souped models. It can be seen that uniform souping of all four models leads to performance regression compared to \textit{MetaMathOctopus-MAPO} and uniform soup with candidate selection reduces this regression. Weight tuning in \soce{} improves the final performance by 1.57\% relative compared to the best baseline model.}
\label{mgsm-70b-sose}
\end{minipage}
\hfill
\begin{minipage}{0.48\linewidth}
\small
\centering
\begin{tabularx}{\linewidth}{|X|l|}
    \hline
    \multicolumn{1}{|c|}{\textbf{Model}} & \makecell{$\infty$\textbf{Bench}\\\textbf{Accuracy}} \\
    \hline
    Model Candidate 1 & 27.24\% \\
    Model Candidate 2 & 24.87\% \\
    Model Candidate 3 & 26.72\% \\
    Model Candidate 4 & 27.24\% \\
    Model Candidate 5 & \underline{27.44}\% \\
    \hline
    Uniform Soup & 27.44\%\\
    \hline
    Uniform Soup w/ \soce{} Model Selection & 27.85\% \\
    \hline
    \soce{}(Proposed Method) & \textbf{28.0\%} \\
    \hline
\end{tabularx}
\captionof{table}{$\infty$-Bench accuracy of LLAMA 3 70B architecture based models. It can be seen that model selection (corresponding to uniform souping with \soce{} Model Selection row) and weight tuning (corresponding to \soce{} row) leads to improvements. Overall, \soce \space improves the best models candidate by 0.56\%.}
\label{table:infitity_bench_ablation}
\end{minipage}
\end{table} 

\subsection{Large Scale Model Analysis: BFCL, MGSM, and $\infty$Bench}
\label{subsubsec:large-scale-analysis}

\begin{figure*}[htbp]
    \centering
    \begin{subfigure}[t]{0.32\textwidth}
        \centering
        \includegraphics[width=\textwidth, height=1.5\linewidth, keepaspectratio]{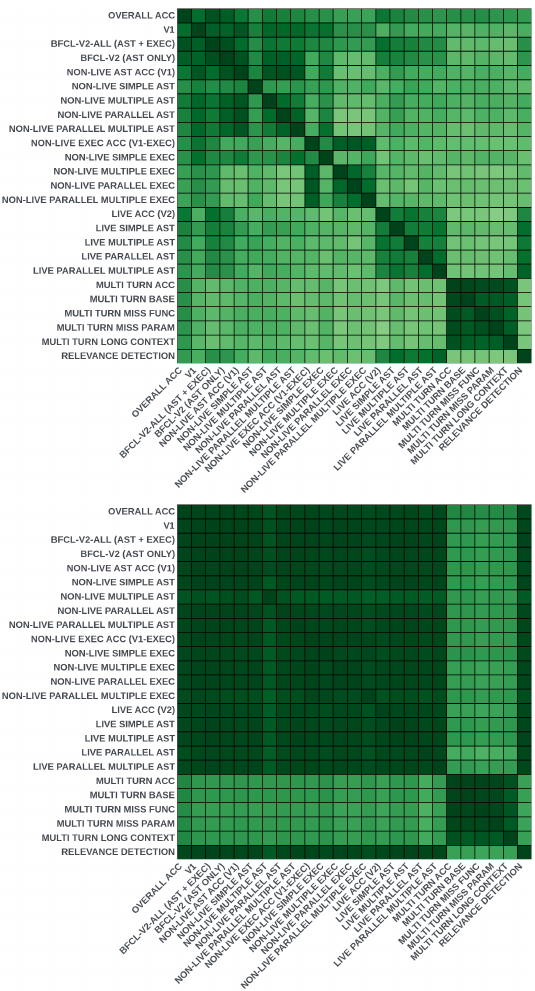}
        \caption{BFCL}
        \label{fig:pearson_corr_a}
    \end{subfigure}
    \hfill
    \begin{subfigure}[t]{0.32\textwidth}
        \centering
        \includegraphics[width=\textwidth, height=1.5\linewidth, keepaspectratio]{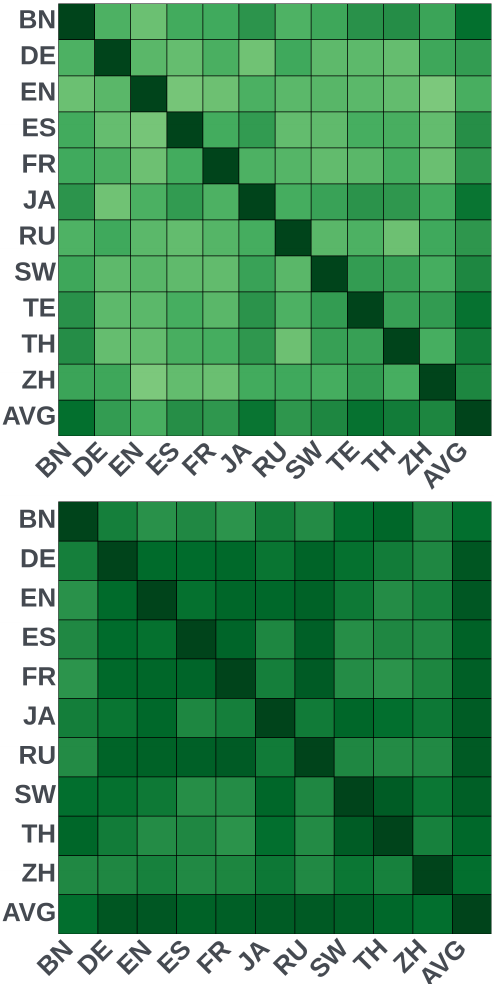}
        \caption{MGSM}
        \label{fig:pearson_corr_b}
    \end{subfigure}
    \hfill
    \begin{subfigure}[t]{0.32\textwidth}
        \centering
        \includegraphics[width=\textwidth, height=1.5\linewidth, keepaspectratio]{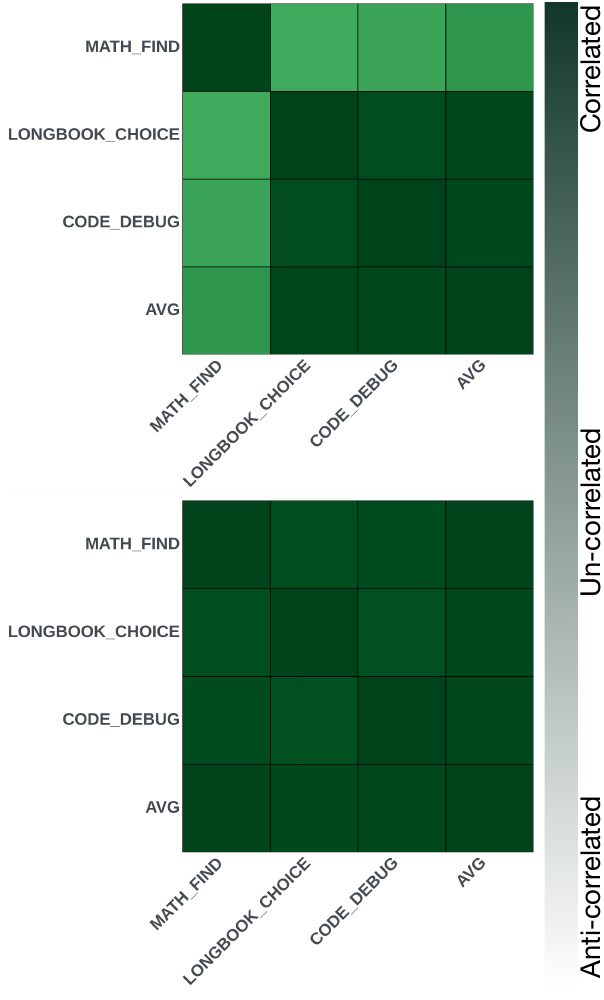}
        \caption{$\infty$Bench}
        \label{fig:pearson_corr_c}
    \end{subfigure}
    \caption{\textbf{Intra-benchmark performance Pearson Correlation (Pre-, and Post- souping)}: This is the pearson correlation matrix of metrics across different categories on (L-R: ) $\sim$800 souped (above) and unsouped (below) checkpoints on BFCL, 40 checkpoints on MGSM and 80 checkpoints for $\infty$Bench. We observe that after souping, the performance across all the categories become highly linearly correlated.}
\label{fig:pearson_correlation3benchmarks}
\end{figure*}
We conducted model souping and evaluation experiments on a vast set of candidate checkpoints across the MGSM, BFCL, and $\infty$Bench benchmarks to systematically investigate the effects of souping. Our analysis yielded the following key findings:
\begin{itemize}
    \item \textbf{Increased Linear Correlation Across Categories after Souping}: Following souping, the performance metrics across different categories exhibit a marked increase in linear correlation. As illustrated in Figure~\ref{fig:pearson_correlation3benchmarks}, the top row represents the Pearson correlation coefficients of checkpoint performance prior to souping, while the bottom row represents the correlations post-souping for BFCL, MGSM, and $\infty$Bench, respectively. These results indicate that souping leads to a more consistent and correlated performance profile across categories within a benchmark.
    \item \textbf{Consistent Performance Gains Across categories}:
    We observe higher average gains across the majority of categories and souping experiments. For example, in case of BFCL, the trained $\sim 500$ unsouped checkpoints had a mean accuracy of 49.85\% (standard deviation of 4.7\%) and randomly souped ($\sim 100$) checkpoints had mean accuracy of 52.75\% (standard deviation of 3.9\%). We dig deeper into this aspect in section \ref{flores_appendix}.
\end{itemize}
The implications of these findings are twofold. First, the process of training large models is often ad hoc, with optimal performance across diverse capabilities typically achieved through extensive experimentation with model parameters and data recipes—an approach that is resource-intensive. Our results suggest that, within a given benchmark, cooperative gains can be achieved in a more deterministic and systematic manner by leveraging strong baselines and principled souping. Second, the observed overall improvements in the average metric values across souping candidates provide strong evidence of robustness of models candidates by adopting end-to-end souping methodologies.

\subsection{\soce{} Candidate selection experiments}
\begin{figure*}[ht]
    \centering
    \begin{subfigure}[b]{0.48\textwidth}
        \includegraphics[width=\textwidth, height=0.6\textheight, keepaspectratio]{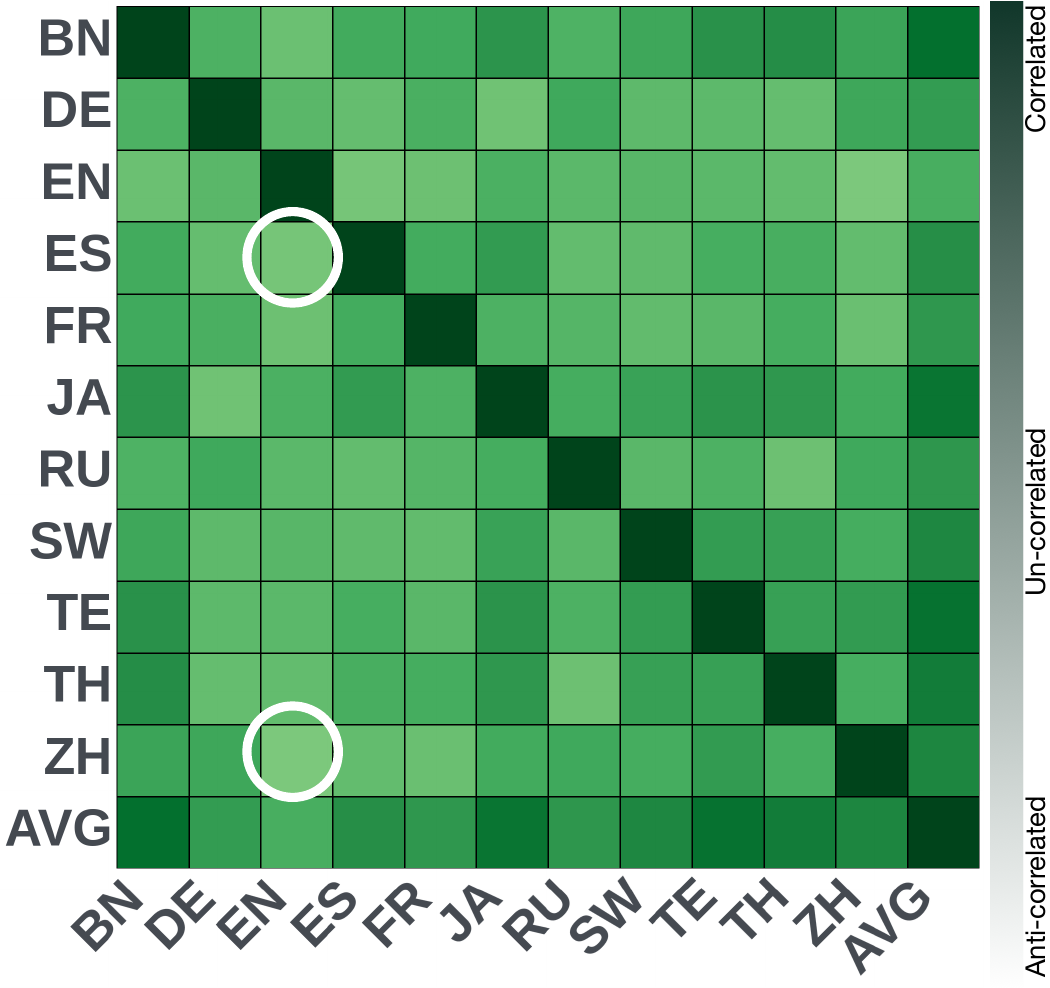}
        \caption{}
        \label{fig:mgsm_analysis_a}
    \end{subfigure}
    \hfill
    \begin{subfigure}[b]{0.48\textwidth}
        \begin{minipage}[t]{\textwidth}
            \includegraphics[width=\textwidth, height=0.29\textheight, keepaspectratio]{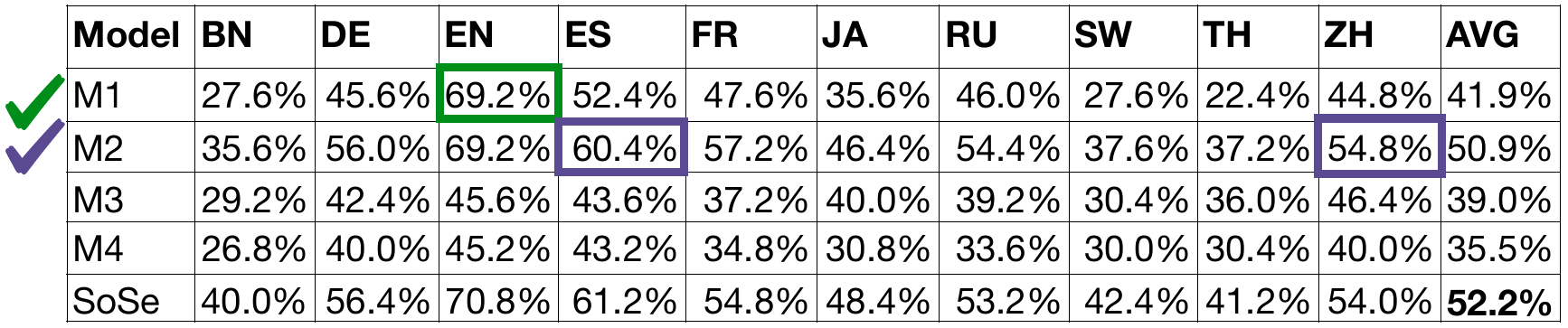}
            \caption{}
            \label{fig:mgsm_analysis_b}
        \end{minipage}
        \begin{minipage}[b]{\textwidth}
            \includegraphics[width=\textwidth, height=0.28\textheight, keepaspectratio]{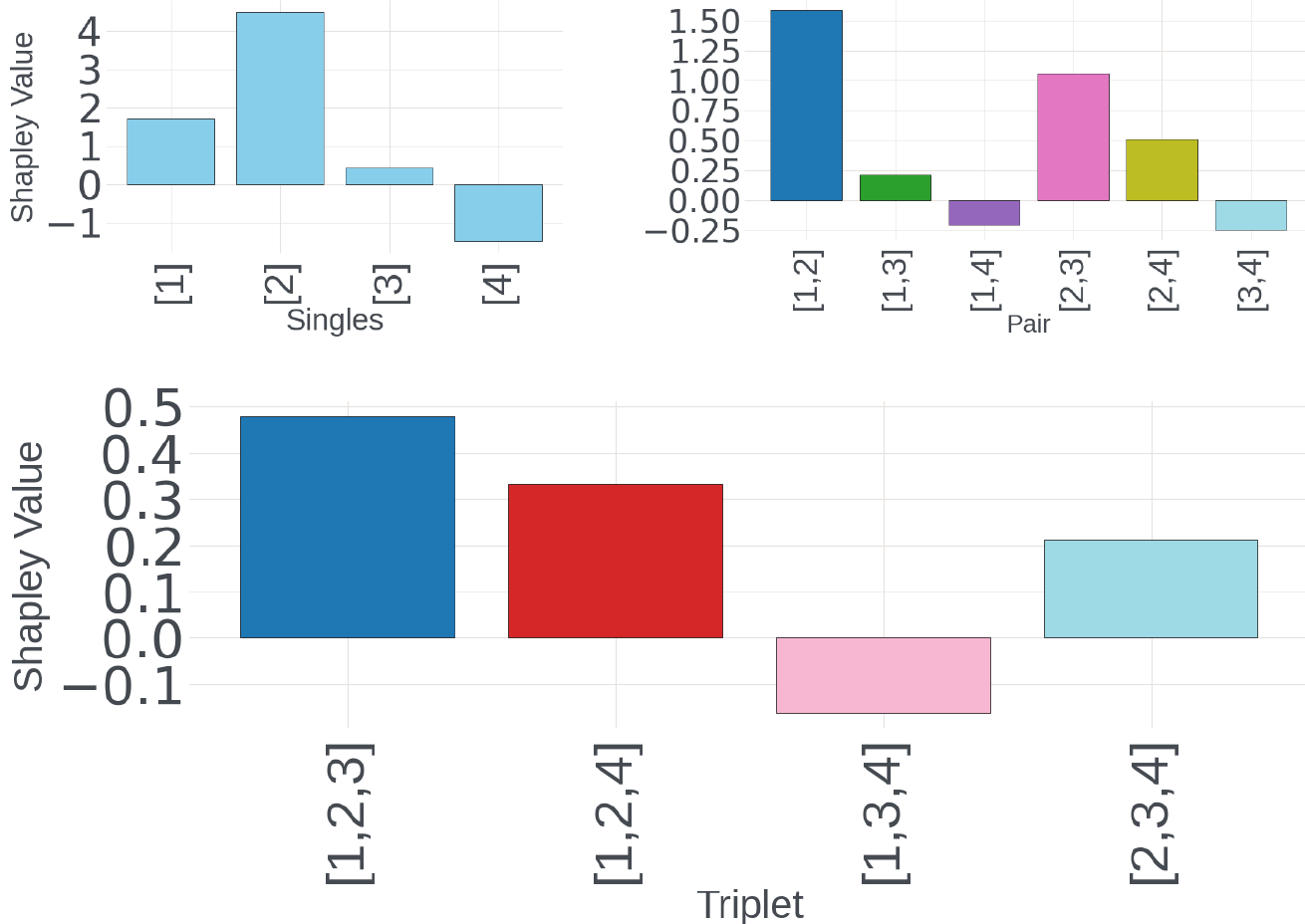}
            \caption{}
            \label{fig:mgsm_analysis_c}
        \end{minipage}
    \end{subfigure}
     \caption{\textbf{Shapley value Analysis}: Figure (a) displays the linear correlation amongst categories of the MGSM benchmarks across 80 checkpoints. Table (b) shows the performance per MGSM benchmark category for a set of 4 finetuned huggingface candidate models, and Figure (c) shows the Shapley value plots for single, pairs and triplets of candidates. We clearly see that M1 and M2 are the experts for the least-correlated categories (ES-EN and ZH-EN) and they are also the strongest contributor pair. In parallel, M1 is a strong parent and M4 is a weak parent and the shapley values reflect that as well showcasing that the strength of \soce's candidate selection approach. }
    \label{fig:mgsm_shapely}
\end{figure*}

\soce\ comprises two components: (1) leveraging correlation patterns to identify experts for uncorrelated categories, and (2) assigning appropriate weights to these experts. We analyze the significance of checkpoint diversity and the impact of anti-correlations within benchmarks.
On MGSM and BFCL, \soce\ yields substantial improvements where distinct experts can be identified across anti-correlated checkpoints (Figures~\ref{fig:mgsm_analysis_a},~\ref{fig:mgsm_analysis_b}). Where clear experts are harder to identify—such as FLORES-36 (Table~\ref{tab:bleu-scores-subbench})—or where benchmarks exhibit minimal anti-correlation, gains are more modest (Table~\ref{table:infitity_bench_ablation}, Figure~\ref{fig:pearson_corr_c}).

\section{Discussion}
In the general case, the \soce{} can be further explored as a broader solution for adding new domain or capabilities to existing open models, overcoming overfitting (Table \ref{tab:overfit-results}). We anticipate future research covering:

\textbf{Application to many more tasks}: \soce{} can be widely applied to the multi-task training objectives, including 
\begin{itemize}
    \item multilingual applications: a task-specific checkpoint + a language specific checkpoint merged (like in \cite{Akiba_2025});
    \item application to combine capabilities at category level instead of capability level like in previous works \cite{cohere2025commandaenterprisereadylarge};
    \item enabling specific use cases, where the training data should remain private, but the checkpoint and its unique capability can be spread across the model family by souping.
\end{itemize}

\textbf{Societal Impact}: While the yearly increase in computational scale leads to growing inequality in academic access to frontier models (see the detailed analysis in \cite{kogkalidis2024tablesnumbersnumbers}), the presented method proposes both easily accessible low cost opportunities for the broader community efforts and also promotes the iterative reuse of the existing pretrained models, which can lead to significant savings in computing resources.

This framework alone can significantly expand the collaboration opportunities in the open source landscape: LLama-derivatives model family currently approaches 150k models\footnote{\url{https://huggingface.co/models?search=llama}} with groups of the same size and architecture that can be souped.

\section{Conclusion}

This work introduces \soce, a category-aware, principled model souping technique that achieves state-of-the-art results and performance improvement across diverse domains. We demonstrate that this approach not only improves overall benchmark performance but also enhances consistency and robustness across categories. We hope these findings inspire further research into efficient model aggregation and reuse within the community. We believe that the presented method can propose a new perspective for the OSS community to reuse and revive derivative models of the same architecture. 
\section{Acknowledgments}  
\label{sec:ack}
We would like to express our sincere gratitude to Andrew Budker and Ricardo Silveira Cabral for their support throughout this work.
We also thank Shishir Patil for his insightful advice on the Berkeley Function Calling Leaderboard evaluations.


\section*{Limitations}
\label{sec:limitations}
\subsection*{General Method Limitations}
The results presented in this paper advance our understanding of model souping, but some methodological and conceptual limitations warrant more discussion.

\textbf{Benchmark structure}: Our method assumes that the target benchmark ships with an explicit sub-categorical decomposition and that each sub-category contains enough data points for per-category scores to be estimated reliably across a sufficient number of models. All experiments in this paper use benchmarks with pre-defined category labels. Many benchmarks do not expose such structure, and we leave the discovery of \emph{implicit} benchmark structure (for example, by means of clustering) to future work.

\textbf{Evaluation Protocol and Test-Score Reliance:} \soce{} performs candidate selection and weight optimization entirely in benchmark performance space: it consumes the per-subcategory scores that public leaderboards already report for existing models, both to identify weakly correlated capability clusters and to grid-search soup weights. Our intended protocol is to derive this correlation structure and these weights from a benchmark's \emph{validation} split and to reserve the test split for final reporting, operating under the standard ML assumption that the two splits are drawn from the same or similar distribution. None of the benchmarks considered here provides such splits, hence we relied on the direct leaderboard scores to pick candidates. In order to understand the risk of overfitting to the BFCL benchmark given test score reliance, we evaluated the merged models on several unrelated capabilities and benchmarks and report these results in  Table \ref{tab:overfit-results}. The souped models show no significant  regression while hill-climbing on BFCL, indicating that the correlation structure captured reflects genuine model properties rather than test-set artifacts. We encourage the LLM Evaluation community to release explicit validation and test splits, so that selection, tuning, and hill-climbing need not be performed directly on the test set which has been a concern that extends well beyond model merging to fine-tuning and LLM-training generally.

\textbf{Application in the model training practice}: In this work, we have only tested the souping of the 'final' post-trained and aligned checkpoints. As other works have shown, the models can be souped after pre-training \cite{li2025modelmergingpretraininglarge}, after post-training \cite{wortsman2022model}, with adapters\footnote{\url{https://huggingface.co/docs/peft/en/developer_guides/model_merging}}.  
We do not recommend souping of the checkpoints from different training stages, as well as souping unaligned or uncensored models with the aligned ones to avoid the inheritance of risks. 

We also want to highlight that all the experiments are carried out on Llama 3~\citep{dubey2024llama} derivative models and essentially have the same pre-trained checkpoint. We are currently unaware if souping requires the same pre-trained checkpoint or can it work with different pre-trained checkpoints as well.

\subsection*{Scaling Laws}
While model souping has demonstrated promising results, it is important to consider potential limitations and diminishing returns as more models are combined. The extent to which performance continues to improve may depend on the diversity and capabilities of the individual models being souped. We have not yet systematically tested the upper bounds of this approach, and it is possible that there exists an optimal strategy for souping checkpoints that varies according to the capability differences between the models involved. Further empirical investigation is needed to better understand these scaling dynamics.

\section*{Ethical Considerations} 
It is important to note that models undergoing souping procedures should be released with the disclosure and detailed description of the said procedures. Otherwise, mechanistic interpretability research on such checkpoints can be compromised: works like \cite{sun2024massiveactivationslargelanguage, yu2025superweightlargelanguage} would show blurred results of an unknown nature. Furthermore, souped models should be evaluated for safety and biases to make sure souping has not injected new biases or amplified existing ones.

\nocite{}
\clearpage
\newpage
\bibliographystyle{plainnat}
\bibliography{custom}

\newpage
\beginappendix
\label{sec:appendix}
\section{Additional experiments}
\subsection{FLORES-36}
\label{flores_appendix}
\begin{figure*}[htbp]
    \centering
    \includegraphics[width=\linewidth]{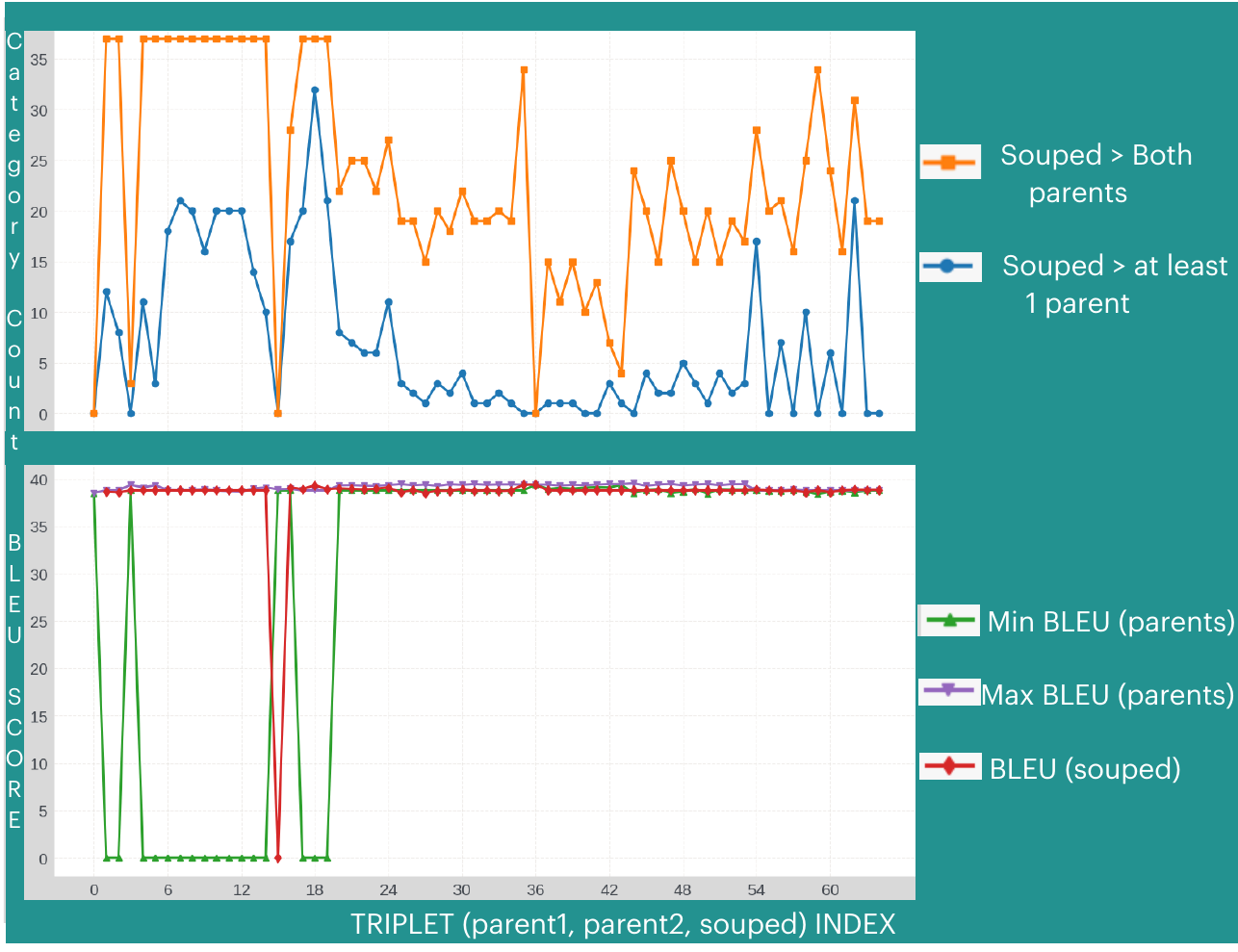}
    \caption{\textbf{Analysis of performance of 37 souped and unsouped checkpoint on Flores-36)}: The x-axis contains the index of one souping triplet, i.e, the two parents and the souped output on the FLORES-36 benchmark. The y-axis in the top figure is the count of the number of categories in FLORES-36 and in the bottom figure, it the BLEU metric score. The orange line maps how many souped outputs have a higher score than the at least one parent per category, the blue line maps how many souped outputs have a higher score than both the souped candidates in the top figure. In the bottom figure, the green line shows the smaller average BLEU score between the parents, the purple line shows the higher BLEU score and the red line shows the souped candidate score. }
    \label{fig:flore37_subbench_analysis}
\end{figure*}

FLORES-101~\citep{flores101_goyal2021}: Measures translation quality and multilingual understanding across a diverse set of language pairs. We use a subset containing translations in 18 languages to and from English. These are only used for ablation studies. We refer to this subset at FLORES-36. We provide a deep dive on FLORES-36 on the model stability and avg increase via souping. We randomly souped (in pairs) 37 checkpoints fine-tuned on Llama-70B, and for 35 out of 37 souping experiments, souped candidates had a higher metric score for more than 20 categories (out of 36) with a net positive gain observed across all categories (Figure \ref{fig:flore37_subbench_analysis}). We also present an ablation on FLORES-36 in a similar setting as $\infty$Bench and the results are presented in Table \ref{tab:bleu-scores-subbench}.

\section{Comparison of \soce{} with TIES and DARE approaches}
\label{70b_dare_ties_section}
In this subsection, we state the results for \soce{} compared with TIES and DARE baselines. We presented the 8b results in Section \ref{numerical_results}. In this section, we present the same set of results for 70 billion parameter models. Table \ref{ties_dare_other_benchmarks_70b} shows results on 70b on these benchmarks. We observe similar regressions in performance across IFEval, MMLU-Pro and LongBench as observed for 8b models. While Truth Q/A doesn't show gains compared to all individual ingredient models, the final DARE/TIES models are better than 2 out 3 ingredients.  

\section{Analysis of TIES and DARE Assumptions}

We conduct a deeper empirical investigation into the underlying assumptions of TIES~\cite{ties_yadav2023tiesmerging} and DARE~\cite{yu2024language}. Our analysis reveals that several core assumptions fail to hold in practice, and that these failures directly explain the performance degradations observed in Tables~\ref{ties_dare_other_benchmarks_8b}, \ref{ties_dare_other_benchmarks_70b}, and \ref{tab:ties_dare_other_benchmarks_bfcl_per_cat_comp}.

\begin{itemize}

    \item \textbf{Near-orthogonal task vectors.} DARE assumes that task vectors derived from models sharing the same base are sufficiently aligned such that magnitude-based pruning preserves meaningful signal. However, pairwise cosine similarity between flattened task vectors averaged $0.042$, indicating near-orthogonality. Pruning entries from such unaligned vectors destroys shared signal rather than isolating task-specific components.

    \item \textbf{Majority sign conflicts.} TIES resolves destructive interference via majority-vote sign election. We find that $52.99\%$ of active parameters exhibited sign conflicts (2-vs-1), forcing TIES to override minority signs for over half of all merged parameters. Furthermore, with a smaller set of candidate models, majority voting does not necessarily correlate with directional information, undermining the validity of this resolution strategy.

    \item \textbf{Extreme conflict among high-magnitude parameters.} TIES retains only top-magnitude parameters under the assumption that these carry consensus signal. We find the opposite: $81.63\%$ of top-$10\%$ magnitude parameters exhibited sign conflicts, meaning the most important parameters are precisely where models disagree most.

    \item \textbf{Parameter droppability.} DARE assumes that a large fraction of task-vector parameters can be dropped---either randomly or by magnitude---without meaningful loss. However, indiscriminate dropping at scale may be catastrophic, as it risks disrupting critical or high-risk neurons~\cite{achillesheel_qin2026}.

\end{itemize}

\noindent These failures manifest clearly in the experimental results. \soce{} consistently outperforms both TIES and DARE on BFCL-v3, with degradation most severe on tasks requiring long-context understanding and deep reasoning, as measured by MMLU-Pro and LongBench. We hypothesise that the sign-conflict resolution in TIES and the rescaling in DARE introduce artificial noise that disrupts the delicate, calibrated weight structure required for tasks involving deeper memory, recall, and extended context. These findings stand in contrast to prior work~\cite{ties_yadav2023tiesmerging,yu2024language} showing TIES and DARE to be superior or on par with simple model averaging. Our results demonstrate that simple averaging with principled candidate selection yields more stable and robust models, while TIES and DARE---despite preserving certain capabilities---introduce severe regressions in domains such as tool use.
\begin{table}[!ht]
\centering
\resizebox{\columnwidth}{!}{%
\begin{tabular}{lcccccc}
\toprule
\textbf{Method} & \textbf{Density} & \textbf{Multi-Turn Acc} & \textbf{AST} & \textbf{Live} & \textbf{Relevance} & \textbf{Irrelevance} \\
\midrule
Baseline (ToolACE) & --     & 35.75\% & 87.77\% & 80.23\% & 72.22\% & 89.94\% \\
Baseline (watt)    & --     & 39.50\% & 86.67\% & 76.10\% & 77.78\% & 75.81\% \\
Baseline (xLAM)    & --     & 69.50\% & 84.58\% & 66.46\% & 77.78\% & 62.69\% \\
\midrule
DARE-Linear        & 0.3    & 0.00\%  & 79.40\% & 54.20\% & 72.22\% & 74.67\% \\
DARE-TIES          & 0.5    & 0.00\%  & 71.56\% & 57.93\% & 44.44\% & 78.26\% \\
TIES               & 0.2    & 0.00\%  & 75.67\% & 55.80\% & 77.78\% & 74.42\% \\
\midrule
\textbf{SoCE (Proposed)} & -- & 70.00\% & 84.85\% & 74.32\% & 77.78\% & \textbf{77.83\%} \\
\bottomrule
\end{tabular}
}
\caption{Comparison of merging methods across BFCL evaluation metrics}
\label{tab:ties_dare_other_benchmarks_bfcl_per_cat_comp}
\end{table}

\begin{table*}[htbp]
\centering
\resizebox{\textwidth}{!}{%
\begin{tabular}{llllll}
\toprule
\textbf{Property} & \makecell[l]{\textbf{SoCE}\\\textbf{(Proposed Approach)}}  & \makecell[l]{\textbf{Task Arithmetic\cite{task_arithmetic_ilharco2023editing}}\\\textbf{(TA)}} & \makecell[l]{\textbf{RegMean}\\\cite{jin2023dataless_regmean}} & \makecell[l]{\textbf{Slerp + TAT}\\\cite{slerp_jang2024}} & \makecell[l]{\textbf{CommandA}\\\cite{cohere2025commandaenterprisereadylarge}} \\
\midrule
\textbf{Operates on} 
    & Raw LLM weights 
    & Task vectors 
    & Raw weights 
    & Embeddings 
    & Raw LLM weights \\
\addlinespace
\textbf{Candidate selection} 
    & \makecell[l]{Yes\\(correlation-guided)} 
    & No 
    & No 
    & No
    & No (domain-designed) \\
\addlinespace
\textbf{Weight optimisation} 
    & \makecell[l]{Non-uniform\\grid search} 
    & \makecell[l]{Scalar $\lambda$,\\grid search} 
    & \makecell[l]{Closed-form\\($X^\top X$)} 
    & \makecell[l]{Scalar $\alpha$,\\manual} 
    & \makecell[l]{Heuristic + perturbation search} \\
\addlinespace
\textbf{Data needed at merge} 
    & \makecell[l]{Benchmark\\eval scores} 
    & None 
    & $X^\top X$ matrices 
    & None (Slerp) 
    & Eval scores (${\sim}500$ metrics)  \\
\addlinespace
\textbf{Modifies parameters} 
    & \makecell[l]{Yes\\(weighted avg)} 
    & Yes (delta ops) 
    & \makecell[l]{Yes\\(reweighted avg)} 
    & \makecell[l]{No\\(embedding interp)} 
    & \makecell[l]{Yes (weighted avg)}  \\
\addlinespace
\textbf{Training required} 
    & No 
    & No 
    & No 
    & \makecell[l]{Yes\\(TAT, lightweight)}
    & \makecell[l]{Yes (purpose-built experts) }  \\
\addlinespace
\textbf{Interference handling} 
    & \makecell[l]{Complementary\\expert selection} 
    & None explicit 
    & \makecell[l]{Via $X^\top X$\\reweighting} 
    & N/A 
    & Consistency regularisation \\
\addlinespace
\textbf{Formal guarantee} 
    & No 
    & No 
    & \makecell[l]{Yes\\(linear models only)} 
    & \makecell[l]{Yes\\(Slerp geometry only)} 
    & No\\
\addlinespace
\textbf{Scalability with $N$} 
    & \makecell[l]{Scales\\(selects subset)} 
    & Degrades 
    & \makecell[l]{Scales\\(closed-form)} 
    & \makecell[l]{N/A\\(2-component)} 
    & \makecell[l]{Scales (7 experts; eval bottleneck)}  \\
\addlinespace
\makecell[l]{\textbf{Risk to critical neurons}\\\cite{achillesheel_qin2026}} 
    & \makecell[l]{Low\\(conservative avg)} 
    & Moderate 
    & Low--moderate 
    & \makecell[l]{Very low\\(TAT only)} 
    & Low (conservative avg) \\
\addlinespace
\textbf{Models needed} 
    & Subset of candidate pool 
    & $\geq 2$ 
    & $\geq 2$ 
    & 1 + LoRA 
    & All trained experts (7) \\
\addlinespace
\textbf{Domain} 
    & (V)LLM (Complex benchmarks) 
    & Vision + NLP 
    & NLP 
    & Vision-language retrieval 
    & LLM (enterprise, 111B)   \\
\addlinespace
\textbf{Pretrained model used} 
    & No 
    & As origin (implicit) 
    & No 
    & No 
    & Yes (shared instruct init.) \\
\addlinespace
\textbf{Category structure} 
    & \textbf{by within-benchmark} 
    & diverse tasks
    & diverse tasks 
    & retrieval 
    & a broad enterprise  \\
\addlinespace
\bottomrule
\end{tabular}%
}
\caption{\textbf{Comparison of \soce{} against other Model Merging methods}, showcasing its utility - operating in the parameter space, depending on publicly available evaluation protocols and selecting and souping candidates in a way that's tractable and elegant while improving performance in the LLM space. }
\label{tab:rw-comparison-extended}
\end{table*}

\begin{table}[t]
\centering
\label{tab:70b_results}
\resizebox{0.85\columnwidth}{!}{%
\begin{tabular}{ll ccccc}
\toprule
\textbf{Method} & \textbf{$d$} & \textbf{TQA} & \textbf{IFEval} & \textbf{MMLU-Pro} & \textbf{LongBench} & \textbf{BFCL} \\
\midrule
xLAM-70B & — & 55.7 & 65.3 & 53.6 & 41.53 & 78.45 \\
CoALM-70B & — & 51.6 & 67.4 & 41.6 & 36.76 & 54.49 \\
watt-70B & — & \textbf{60.4} & 73.6 & \textbf{61.2} & 43.16 & 73.57 \\
\midrule
TIES & 0.7 & 57.1 & 58.4 & 7.5 & 2.11 & 46.5 \\
DARE-Linear & 0.9 & 59.1 & 66.4 & 6.3 & 2.75 & 46.3 \\
DARE-TIES & 0.9 & 57.6 & 61.0 & 8.1 & 2.09 & 46.2 \\
\midrule
\soce{} & — & 58.12 & \textbf{74.0} & 58.94 & \textbf{45.62} & \textbf{80.68} \\
\bottomrule
\end{tabular}}
\caption{Performance of Merging Methods on 70B models: Comparison of \soce{} with TIES and DARE methods across Truthful QA (TQA), IFEval, MMLU-Pro and LongBench. All metrics are reported in percentages.}
\label{ties_dare_other_benchmarks_70b}
\end{table}

\section{Understanding Impact on Other Benchmarks while souping for one benchmark}

We evaluated \soce{} models that had state-of-the-art performance on BFCL-v3 on the following unrelated benchmarks to probe for overfitting to BFCL:  
\begin{enumerate}[noitemsep, topsep=0pt, leftmargin=*]
    \item Hellaswag~\cite{zellers2019hellaswag} (commonsense reasoning via sentence completion)
    \item IFEval~\cite{zhou2023instruction} (instruction following benchmark)
    \item BIG-Bench Hard~\cite{suzgun2023challenging} (BIG-Bench~\cite{srivastava2023beyond} tasks that were hard for humans across reasoning and deduction)
\end{enumerate}
We present these results in Table~\ref{tab:overfit-results} for both 8 billion (8B) parameters model and 70 billion (70B) parameters model. We observed the \soce{} models are either best (e.g. 70B model on IF-Eval) or second best (e.g. 8B on IF-Eval) in many cases and we observe no significant regressions (maximum regression of \soce{} models compared to the best candidate is 1.08\% for IF-Eval 8B \soce{} model). This shows a strong evidence that \soce{} does not lead to overfitting on the benchmark its tuned on, even with the test-score reliance as discussed in limitations section.


\begin{table*}[!htbp]
\centering

\begin{tabularx}{\textwidth}{|l|X|X|X|X|}
\hline
\textbf{Models} & \textbf{BFCL-v3} & \textbf{Hellaswag} & \textbf{IFEval} & \textbf{BBH} \\
\hline
ToolACE-2-8B & 68.73\% & 78.92\% & 51.68\% & 44.80\% \\
xLAM-2-8b-fc-r & 72.37\% & 77.57\% & 45.00\% & 63.10\% \\
watt-tool-8B (FC) & 67.79\% & 78.59\% & 50.36\% & 36.54\% \\
SoCE (Proposed Method) & \textbf{76.50\%} & 78.37\% & 50.60\% & 63.06\% \\
\hline
\end{tabularx}

\vspace{0.4em}
{\small (a) 8B Souping}

\vspace{1em}

\begin{tabularx}{\textwidth}{|l|X|X|X|X|}
\hline
\textbf{Models} & \textbf{BFCL-v3} & \textbf{Hellaswag} & \textbf{IFEval} & \textbf{BBH} \\
\hline
xLAM-2-70b & 78.45\% & 82.80\% & 65.23\% & 82.46\% \\
watt-tool-70B & 73.57\% & 84.35\% & 73.62\% & 86.22\% \\
CoALM-70B & 54.49\% & 83.96\% & 67.39\% & 83.72\% \\
SoCE (Proposed Method) & \textbf{80.68\%} & 83.91\% & 73.98\% & 85.72\% \\
\hline
\end{tabularx}

\vspace{0.4em}
{\small (b) 70B Souping}

\caption{\textbf{BFCL Souping Overfitting Results:} Performance of state-of-the-art BFCL-v3 \soce{} models on Hellaswag~\cite{zellers2019hellaswag}, IFEval~\cite{zhou2023instruction} and BIG-Bench Hard~\cite{suzgun2023challenging} to understand overfitting to BFCL-v3.}
\label{tab:overfit-results}
\end{table*}

\section{Deep dive on BFCL State-of-the-art Performance: Winrate Analysis of Souped Models and Individual Candidates}

On the BFCL benchmark, for 70B parameters dense models, we analyze the win rate of \soce{} over individual models in the soup: xLAM-2-70b-fc-r ~\citep{prabhakar2025apigen}, CoALM-70B~\citep{acikgoz2025can}, and watt-tool-70B~\cite{wattai_70b}. \\

\noindent We first examine the proportion of tasks solved by \soce{} that were also solved by each individual model. \soce{} successfully completes 97.2\% of the tasks solved by xLAM-2-70b-fc-r~\citep{prabhakar2025apigen}, 97.1\% of those solved by CoALM-70B~\citep{acikgoz2025can}, and 97.1\% of those solved by watt-tool-70B~\cite{wattai_70b}.
This indicates that \soce{} retains most of the capabilities of individual models in the soup.\\

\noindent Next, we assess \soce's win rate on tasks where individual models fail.
When individual models all fail on a given task, \soce{} succeeds in 8.4\% of cases (32 out of 380 tasks), demonstrating \soce's ability to solve new tasks that none of the models in the soup was able to handle.\\

\noindent Further, when one individual model in the soup fails on a given task, and not the others, \soce{} is able to complete the task in 93.0\% of cases, highlighting the robustness of the proposed souping approach.

\section{Sensitivity Analysis of the Correlation Threshold $\tau$ and Computational Upper Bounds of \soce}
\label{tau_appendix}
The sensitivity of \textsc{SoCE} to $\tau$ is bounded by two degenerate
extremes: at $\tau \to -1$ the anti-correlation criterion is never satisfied
and the method reduces to single-model selection; at $\tau \to 1$ it degenerates
to an unconstrained soup over all $B$ subcategory experts, at which point the
grid search reaches its computational upper bound of $\binom{B+9}{9}$ candidate
merges. For the benchmarks in this paper these upper bounds are 2,002
($\infty$-Bench), 5,005 (BFCL), and 167,960 (MGSM). The operative regime of
\textsc{SoCE} --- small or negative $\tau$, $N(\tau) \leq 4$, tractable grid
search --- lies strictly between these extremes. The anti- and de-correlation
selection criterion is not merely a heuristic for complementarity; it is
simultaneously the mechanism that keeps the method computationally tractable
by preventing the expert count from approaching the structural upper bound $B$.

\section{Post-hoc validation of our candidate selection process using Shapley Values }
    

\textbf{Data and Game Theory:} Cooperative game theory, and Shapley value~\citep{shapley:book1952}, has found broad application in machine learning for quantifying component contributions. \citet{ghorbani2019data} introduced Data Shapley, applying the Shapley value framework to equitably value individual training data points based on their marginal contribution to model performance. \citet{rozemberczki2022shapley} provided a comprehensive survey of Shapley value applications across ML, spanning feature attribution, data valuation, model explanation, and federated learning. While alternative power indices such as the Banzhaf value have been proposed for data valuation~\citep{wang2023banzhaf}, the Shapley value remains the canonical choice due to its uniqueness properties: \citet{Dubey1975OnTU} proved that it is the unique allocation satisfying efficiency, symmetry, null player, and additivity-properties that make it a natural choice for analyzing model contributions in coalitional settings. Unlike prior uses of Shapley values in ML, which typically operate at the data-point or feature level, \soce{}'s application is at the level of model checkpoints-a novel use case enabled by the small number of candidates (making exhaustive coalition evaluation tractable). Importantly, the Shapley value analysis in \soce{} helps with post-hoc validation and interpretability rather than optimization, confirming that the selected category experts are indeed the highest-contributing members of the coalition.\\
\noindent \textbf{Cooperative Game Theory and the Shapley Value}:
Cooperative game theory addresses situations where multiple agents collaborate as a team. A (transferable-utility) cooperative game is specified by a set of players
$A = \{a_1, a_2, \ldots, a_n\}$
and a characteristic function
$v: 2^A \rightarrow \mathbb{R},$
which assigns to each coalition \( C \subseteq A \) a real value \( v(C) \) representing the utility or performance achieved by that coalition.
The \textbf{Shapley value}~\cite{shapley:book1952} is a fundamental concept in cooperative game theory, providing a principled way to quantify each player's individual contribution to the team's overall success, while satisfying key fairness properties~\cite{Dubey1975OnTU}.
Formally, for a player \( a_i \in A \) in a game with characteristic function \( v \), the \emph{marginal contribution} of \( a_i \) to a coalition \( C \subseteq A \setminus \{a_i\} \) is given by
$m(a_i, C) = v(C \cup \{a_i\}) - v(C)$
Let \( \Pi \) denote the set of all permutations of the \( n \) players, i.e., each \( \pi \in \Pi \) is a bijection \( \pi: A \rightarrow A \). For a permutation \( \pi \), define \( \mathrm{pred}(a_i, \pi) \) as the set of players preceding \( a_i \) in \( \pi \). The marginal contribution of \( a_i \) in permutation \( \pi \) is
$m(a_i, \pi) = v(\mathrm{pred}(a_i, \pi) \cup \{a_i\}) - v(\mathrm{pred}(a_i, \pi))$
The \textbf{Shapley value} \( \phi_i \) for player \( a_i \) is the average marginal contribution of \( a_i \) over all possible permutations:
$\phi_i = \frac{1}{n!} \sum_{\pi \in \Pi} \left[ v(\mathrm{pred}(a_i, \pi) \cup \{a_i\}) - v(\mathrm{pred}(a_i, \pi)) \right]$
Equivalently, the Shapley value represents the expected increase in team value when \( a_i \) joins the team at a random position in a random ordering of the players.\\
\textbf{Shapley Value Analysis in the Context of Model Souping}:
For the Shapley value analysis, we consider a benchmark \( B \), a metric \( F \), and a set \( V \) of candidate models for souping.
As candidates, we analyze either (1) individual checkpoints, (2) pairs, or (3) triplets, treating each candidate as a player and each subset \( C \subseteq V \) as a coalition. This forms a cooperative game where the characteristic function is defined as
$v: 2^V \rightarrow \mathbb{R},$
mapping each subset of candidates to the performance achieved by souping only those candidates.
For any coalition \( C \subseteq V \), the team performance is given by
$v(C) = \mathrm{Performance}\left(\mathrm{Souped}(C); B, F\right),$
where \( \mathrm{Souped}(C) \) denotes the model obtained by souping the candidates in \( C \), evaluated on benchmark \( B \) using metric \( F \). We view the souped models as participants / agents working together to build the best model, bringing different skills / strengths to the table (as shown by the categories within the benchmarks). By souping models, we allow covering these different skills well, as showcased by an improvement in the overall metric. The Shapley value indicates the relative contributions of the sub-models using souping as the combination function in the context of the skills of the set of all models.

\begin{figure*}[htbp]
    \centering
    \resizebox{0.68\textwidth}{!}{%
    \begin{subfigure}{\linewidth}
        \centering
        \includegraphics[width=0.9\linewidth]{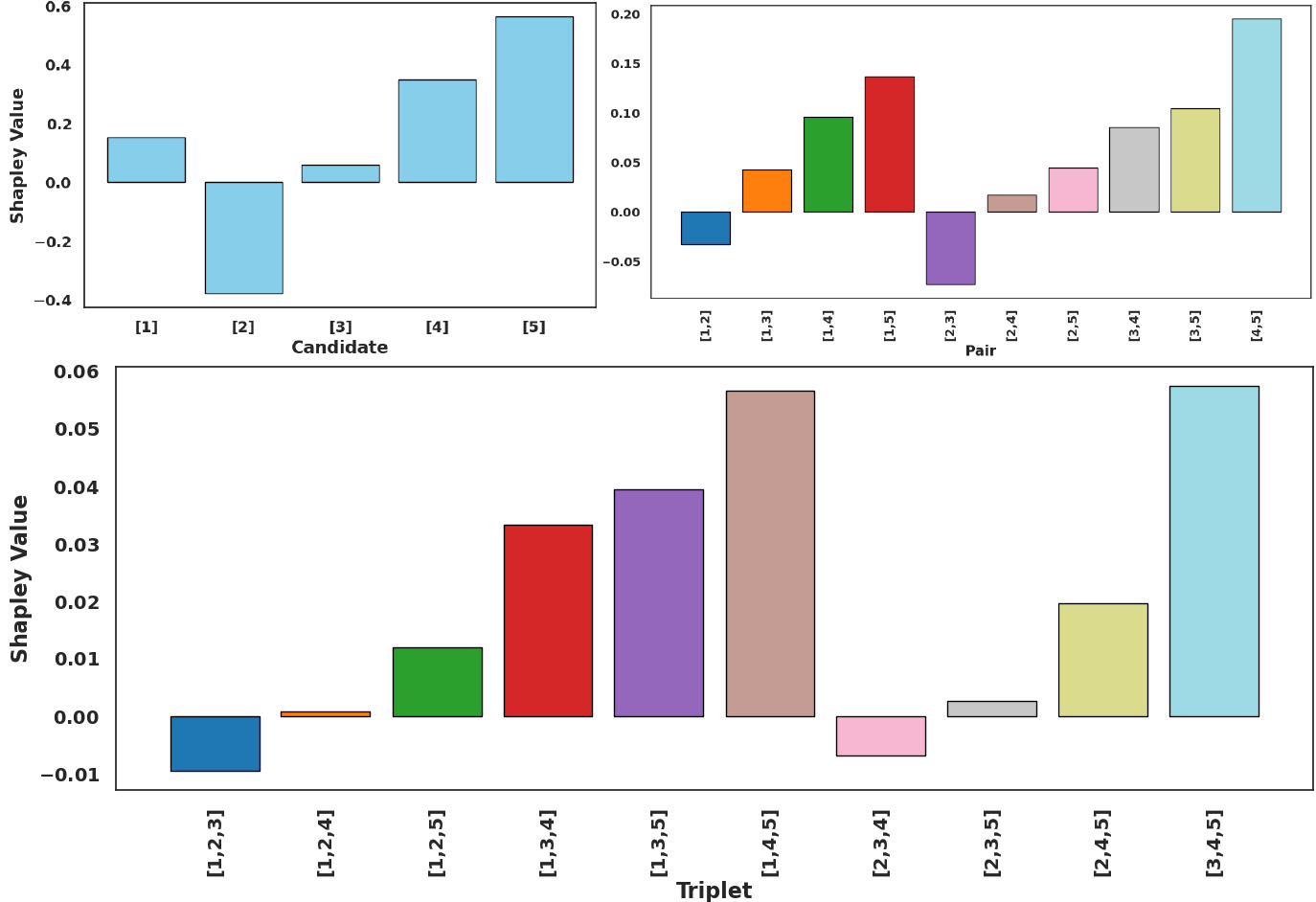}
        \caption{\textbf{$\infty$-Bench for a set of 5 finetuned candidate 70B models.}}
        \label{fig:inf_shapely}
    \end{subfigure}}
    \resizebox{0.68\textwidth}{!}{%
    \begin{subfigure}{\linewidth}
        \centering
        \includegraphics[width=0.9\linewidth]{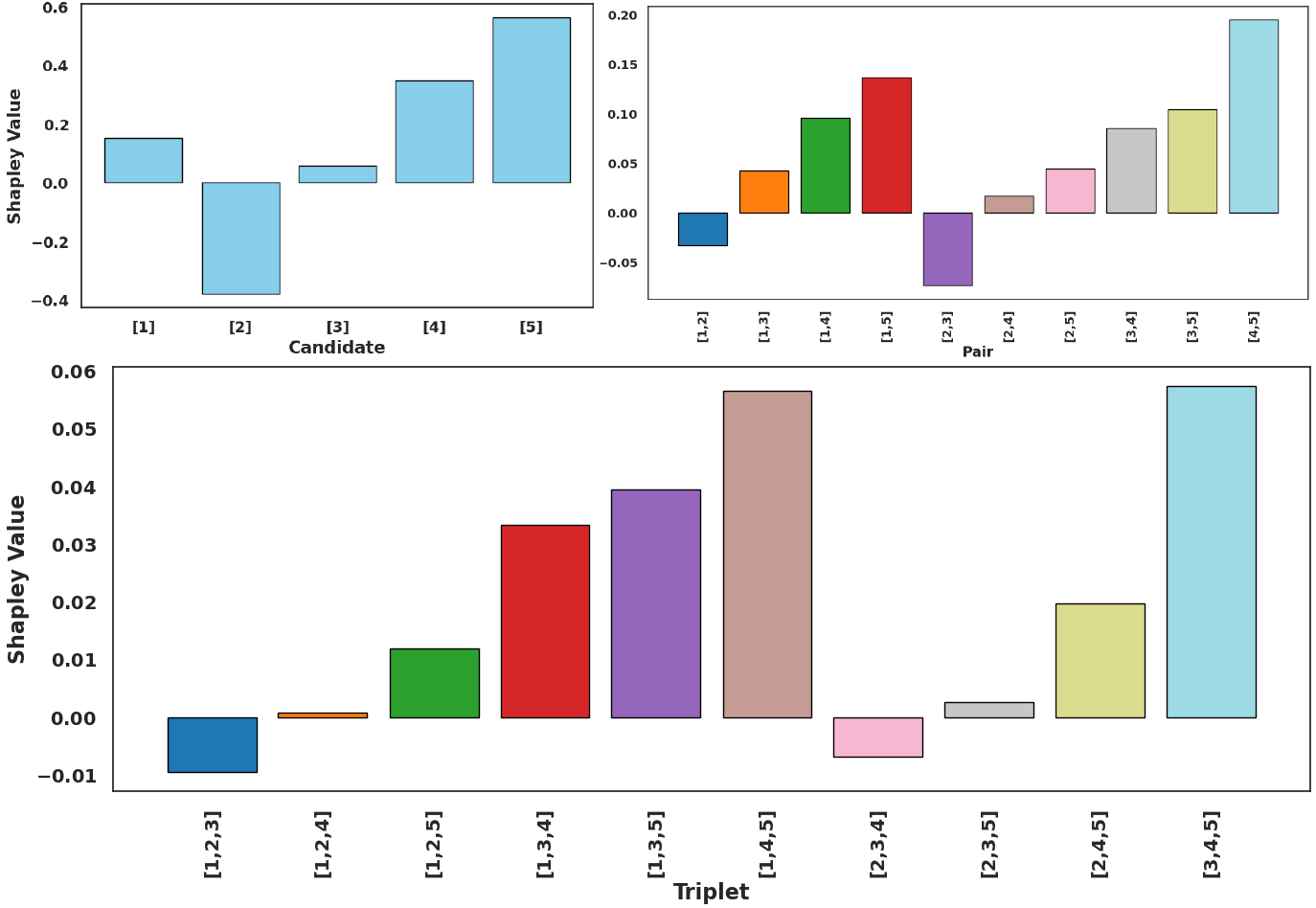}
        \caption{\textbf{MGSM for a set of 4 finetuned huggingface candidate models.}}
        \label{fig:mgsm_shapely}
    \end{subfigure}}
    \resizebox{0.68\textwidth}{!}{%
    \begin{subfigure}{\linewidth}
        \centering
        \includegraphics[width=0.9\linewidth]{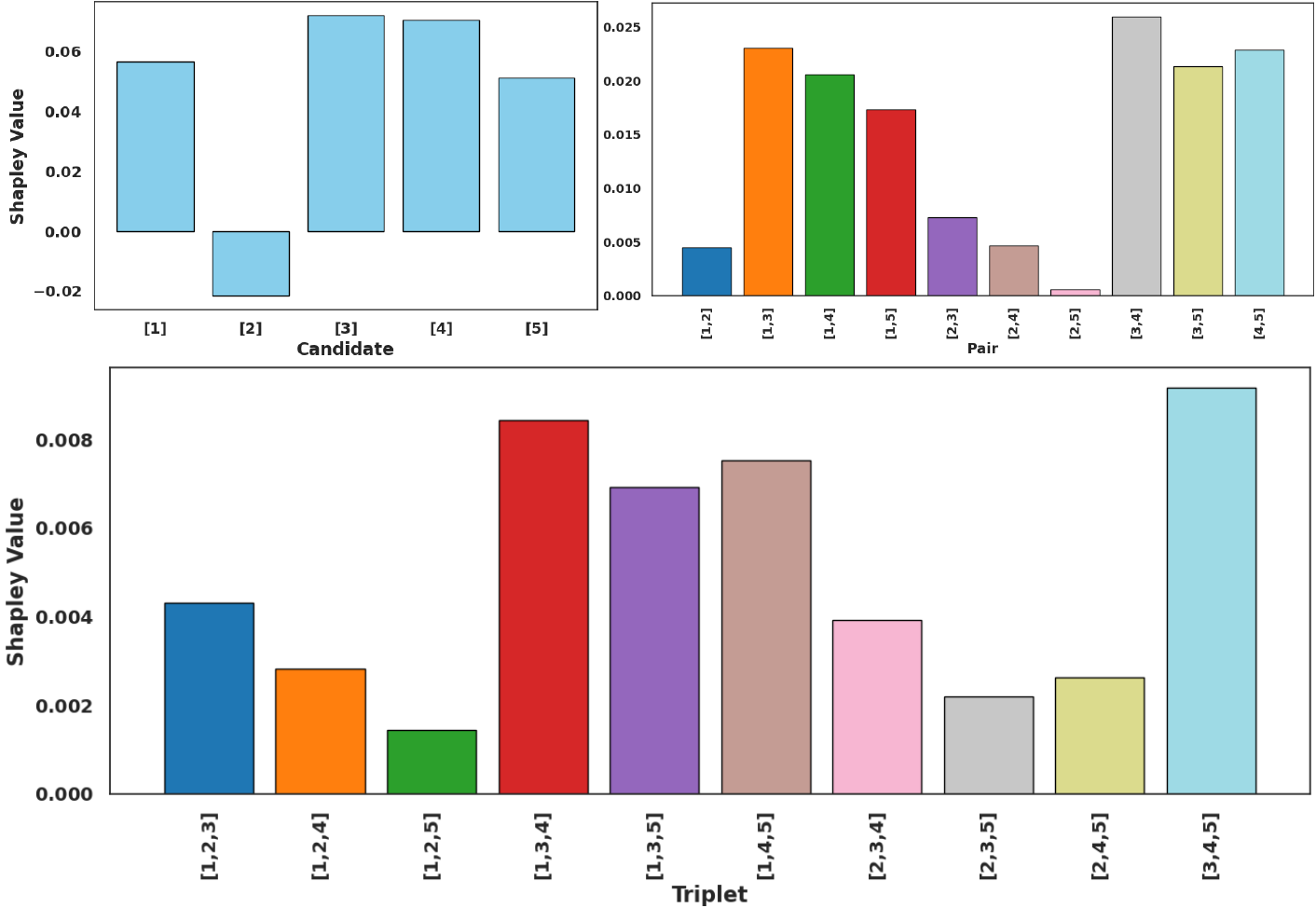}
        \caption{\textbf{FLORES-36-Bench for a set of 5 finetuned candidate 70B models.)}}
    \label{fig:flores_shapely}
        \label{fig:sub3}
    \end{subfigure}}
    \caption{\textbf{Shapley values}: We computed Shapley values~\cite{shapley:book1952} for a small batch of 4-5 models to understand the contributions of model candidates (single, in pairs and in triplets) and their correlation with the \soce \space candidate selection mechanism. We found that \soce's selected candidates often contributed more significantly under the souping paradigm.}
    \label{fig:supp_shapley}
\end{figure*}

\noindent We define our Shapley computation by the following parameters; the benchmark is MGSM, the performance metric is average accuracy, the characteristic function is souping and the set of candidates are (1) 4 open-source (OSS) models finetuned on LLama-7B (2) 6 pairs of the OSS models (3) 4 possible triplets  these for model souping using a set of four candidate models, evaluating all possible combinations.
Separately, we also apply the \soce{} framework to the same set. As shown in Figure~\ref{fig:mgsm_shapely_3fig}, our analysis reveals that model contributions are not uniform; candidates and subsets selected via \soce{} exhibit significantly higher Shapley values. This finding underscores the critical role of \soce’s candidate selection in enhancing ensemble performance.
Additionally, our results indicate that disproportionately weighting a weak parent model can substantially decrease the average score (e.g., assigning weights of (0.05, 0.05, 0.9) on M4 reduces the mean to 37.0). In contrast, consistently including a strong parent model, such as M2 in combination with other candidates, reliably increases the average score. We also extend the same method of analyses to other benchmarks (see Figures~\ref{fig:flores_shapely} and~\ref{fig:inf_shapely}). \\
We computed Shapley values~\cite{shapley:book1952} for a small batch of 4-5 models to understand the contributions of model candidates (single, in pairs and in triplets) and their correlation with the \soce{} candidate selection mechanism. We found that \soce's selected candidates often contributed more significantly under the souping paradigm. We also include a scenario where category experts for uncorrelated categories are not easy to distinguish, i.e. in the case of FLORES-36. In this case, the overall improvements of souping are marginal. This also highlights the advantage of \soce's candidate selection strategy. We present the Shapley values in Figure \ref{fig:supp_shapley}.

\begin{table*}[!t]
\centering
\begin{tabular}{|l|c|c|c|c|}
\hline
\textbf{Model} & \textbf{avg\_$\infty$Bench} & \textbf{inf\_mathfind} & \textbf{inf\_longbookchoice} & \textbf{inf\_codedebug} \\
\hline
1     & $27.24\%$ & $33.71\%$ & \textbf{30.13\%} & $19.80\%$ \\
2     & $24.87\%$ & $32.57\%$ & $23.14\%$ & $19.04\%$ \\
3     & $26.72\%$ & $34.00\%$ & $24.89\%$ & $21.32\%$ \\
4     & $27.24\%$ & $34.00\%$ & $24.89\%$ & $22.59\%$ \\
5     & $27.44\%$ & $32.57\%$ & $27.07\%$ & $23.10\%$ \\
\soce{}  & \textbf{27.85\%} & \textbf{34.57\%} & $27.51\%$ & \textbf{23.60\%} \\
\hline
\end{tabular}
\caption{$\infty$Bench results (percent scale) for various models and combinations.}
\label{tab:infbench-sub}
\end{table*}
\begin{table*}[htbp]
    \centering
    \begin{subtable}[t]{\textwidth}
        \centering
        \resizebox{\textwidth}{!}{%
        \begin{tabular}{|l|c|c|c|c|c|c|c|c|c|}
    \hline
    \textbf{Model} & \textbf{avg\_bleu} & \textbf{eng\_deu} & \textbf{deu\_eng} & \textbf{eng\_spa} & \textbf{spa\_eng} & \textbf{eng\_tha} & \textbf{tha\_eng} & \textbf{eng\_por} & \textbf{por\_eng} \\
    \hline
    1 & 38.97 & 45.60 & 50.18 & 35.49 & 35.49 & 35.29 & 36.33 & 52.96 & 54.79 \\
    2 & 38.74 & 45.76 & 50.03 & 35.77 & 35.77 & 35.48 & 35.32 & 53.01 & 54.41 \\
    3 & 39.07 & 45.85 & \textbf{50.55} & \textbf{35.78} & \textbf{35.78} & 35.82 & 36.43 & 53.12 & 54.75 \\
    4 & 39.04 & 45.95 & 50.09 & 35.35 & 35.35 & 35.89 & 36.16 & 52.85 & 54.61 \\
    5 & 38.94 & 45.75 & 50.38 & 35.15 & 35.15 & 35.59 & 36.10 & 52.59 & 54.42 \\
    \soce & \textbf{39.19} & \textbf{46.22} & 50.39 & 32.06 & 35.72 & \textbf{36.26} & \textbf{36.57} & \textbf{53.27} & \textbf{54.83} \\
    \hline
    \end{tabular}}
    \caption{BLEU scores on FLORES-36 (German, Spanish, Thai, Portuguese.) 70B finetuned models Souping per language translation split. (Part 1 of 4)}
    \end{subtable}
    \hfill
    \begin{subtable}[t]{\textwidth}
        \centering
        
        \resizebox{\textwidth}{!}{\begin{tabular}{|l|c|c|c|c|c|c|c|c|c|}
        \hline
        \textbf{Model} & \textbf{eng\_vie} & \textbf{vie\_eng} & \textbf{eng\_ind} & \textbf{ind\_eng} & \textbf{eng\_hin} & \textbf{hin\_eng} & \textbf{eng\_fra} & \textbf{fra\_eng} & \textbf{eng\_ita} \\
        \hline
        1 & 41.84 & 40.75 & 44.99 & 48.07 & 35.11 & 44.95 & 55.55 & 50.88 & 35.08 \\
        2 & 41.70 & 40.27 & 44.91 & 47.51 & 35.32 & 44.41 & 55.49 & 50.47 & 35.16 \\
        3 & 41.74 & 40.82 & 44.98 & 47.96 & 34.88 & 44.74 & 55.68 & 51.06 & 34.86 \\
        4 & 41.82 & 40.60 & \textbf{45.25} & 48.17 & 34.96 & 44.69 & 55.80 & 50.78 & 35.09 \\
        5 & 41.52 & 40.75 & 44.77 & 48.35 & 34.86 & 44.73 & 55.52 & 50.85 & 34.97 \\
        \soce & \textbf{42.08} & \textbf{41.06} & 45.22 & \textbf{48.50} & \textbf{35.47} & \textbf{45.08} & \textbf{56.15} & \textbf{51.26} & \textbf{35.26} \\
        \hline
        \end{tabular}}
        \caption{BLEU scores on FLORES-36 (Vietnamese, Indonesian, Hindi, French, Italian.) 70B finetuned models Souping per language translation split. (Part 2 of 4)}
    \end{subtable}
    
    \vspace{0.5cm}
    
    \begin{subtable}[t]{\textwidth}
        \centering
        \resizebox{\textwidth}{!}{
        \begin{tabular}{|l|c|c|c|c|c|c|c|c|c|}
        \hline
        \textbf{Model} & \textbf{ita\_eng} & \textbf{eng\_arb} & \textbf{arb\_eng} & \textbf{eng\_ben} & \textbf{ben\_eng} & \textbf{eng\_zhoHs} & \textbf{zhoHs\_eng} & \textbf{eng\_zhoHt} & \textbf{zhoHt\_eng} \\
        \hline
        1 & 38.90 & 34.96 & 45.80 & 29.68 & 37.55 & 33.56 & 34.98 & 17.57 & 31.99 \\
        2 & 39.20 & 35.32 & 45.59 & 29.94 & 37.10 & 33.32 & 34.43 & 17.39 & 31.22 \\
        3 & 38.95 & 35.06 & 46.01 & 29.94 & 37.88 & 33.59 & 34.85 & 18.92 & 31.60 \\
        4 & 38.52 & 35.43 & 45.97 & 30.13 & 37.92 & 33.58 & 34.41 & 21.28 & 31.53 \\
        5 & 38.69 & 35.14 & 46.06 & 29.98 & 37.45 & 33.66 & 34.64 & 18.63 & 31.83 \\
        \soce{} & \textbf{39.31} & \textbf{36.05} & \textbf{46.28} & \textbf{30.31} & \textbf{37.94} & \textbf{33.98} & \textbf{35.06} & 20.34 & 31.89 \\
        \hline
        \end{tabular}}
        \caption{BLEU scores on FLORES-36 (Italian, Arabic, Bengali, Chinese (Simplified/Traditional)) 70B finetuned models Souping per language translation split. (Part 3 of 4)}
    \end{subtable}
    \hfill
    \begin{subtable}[t]{\textwidth}
        \centering
        \resizebox{\textwidth}{!}{
        \begin{tabular}{|l|c|c|c|c|c|c|c|c|c|c|}
        \hline
        \textbf{Model} & \textbf{eng\_jpn} & \textbf{jpn\_eng} & \textbf{eng\_kor} & \textbf{kor\_eng} & \textbf{eng\_rus} & \textbf{rus\_eng} & \textbf{eng\_tur} & \textbf{tur\_eng} & \textbf{eng\_tgl} & \textbf{tgl\_eng} \\
        \hline
        1 & 25.20 & 32.83 & 24.44 & 34.67 & 38.63 & 42.32 & 35.78 & 43.03 & 33.37 & 47.80 \\
        2 & 24.57 & 32.18 & 24.28 & 33.71 & 38.24 & 41.96 & 36.02 & 42.37 & 33.30 & 47.53 \\
        3 & 24.97 & 33.64 & 24.66 & 34.93 & 38.57 & 42.22 & 35.78 & 43.44 & 32.38 & 48.29 \\
        4 & 24.93 & 33.17 & 23.67 & 34.35 & 38.74 & 42.13 & 35.91 & 42.80 & 32.74 & 48.14 \\
        5 & 24.75 & 33.34 & 23.87 & 34.53 & 38.77 & 41.88 & 36.26 & 42.93 & 33.00 & 48.17 \\
        \soce{} & \textbf{25.34} & 33.60 & 24.84 & 34.77 & \textbf{38.88} & \textbf{42.45} & \textbf{36.43} & \textbf{43.26} & \textbf{33.59} & \textbf{48.35} \\
        \hline
        \end{tabular}}
        \caption{BLEU scores on FLORES-36 (Japanese, Korean, Russian, Turkish, Tagalog.) 70B finetuned models Souping per language translation split. (Part 4)}
    \end{subtable}
\caption{\textbf{Per category performance deep dive on FLORES-36 benchmark:} Each language is either translated to (language\_english) or from (english\_language) english. These models are less diverse, i.e, it is difficult to find category experts, and display evidence of marginal gains consistently on average and per sub-benchmark using \soce.}
\label{tab:bleu-scores-subbench}
\end{table*}

\section{Qualitative examples of souping results}
\subsection{Wins when candidates fail}

In tables~\ref{tab:taskjava} and ~\ref{tab:tasksimple}, we show 2 examples of tasks where our proposed approach, \soce, applied to 70 billion parameters dense models, solves problems that none of the individual models in the soup could solve.

\noindent For task simple\_267, \soce \space correctly omits the month parameter, while other models erroneously include month="upcoming".

\noindent For task java\_52, \soce \space correctly uses "Element.class" versus "Element" for the node\_type parameter.

\begin{table*}[ht]
\centering
\begin{tabular}{|l|p{12cm}|}
\hline
\textbf{Task} & \textbf{Description} \\
\hline
java\_52 & In a Java XML processing context, help me obtain a list of all child elements of type \texttt{Element} from a \texttt{Node} representing a SAML assertion \texttt{SAMLAssertionNode}? \\
\hline
\textbf{Model} & \textbf{Output} \\
\hline
xLAM-2-70b-fc-r \textcolor{red}{\(\times\)}& \texttt{[{"name": "SamlObjectSignerTests.getChildren", "arguments": {"node": "SAMLAssertionNode", "node\_type": "Element"}}]} \\
\hline
watt-tool-70B \textcolor{red}{\(\times\)}& \texttt{[SamlObjectSignerTests.getChildren(node='SAMLAssertionNode', node\_type='Element')]} \\
\hline
CoALM-70B \textcolor{red}{\(\times\)}& \texttt{[SamlObjectSignerTests.getChildren(node="SAMLAssertionNode", node\_type="Element")]} \\
\hline
\soce{} \textcolor{green}{\(\checkmark\)} & \texttt{[SamlObjectSignerTests.getChildren(node="SAMLAssertionNode", node\_type="Element.class")]} \\
\hline
\end{tabular}
\caption{Comparison of model outputs for task java\_52 in BFCL v3.}
\label{tab:taskjava}
\end{table*}

\begin{table*}[ht]
\centering
\begin{tabular}{|l|p{12cm}|}
\hline
\textbf{Task} & \textbf{Description} \\
\hline
simple\_267 & Find the top rated modern sculpture exhibition happening in New York in the upcoming month \\
\hline
\textbf{Model} & \textbf{Output} \\
\hline
xLAM-2-70b-fc-r \textcolor{red}{\(\times\)}& \texttt{[{"name": "find\_exhibition", "arguments": {"location": "New York City, NY", "art\_form": "sculpture", "month": "upcoming", "user\_ratings": "high"}}]} \\
\hline
watt-tool-70B \textcolor{red}{\(\times\)}& \texttt{[find\_exhibition(location="New York City, NY", art\_form="sculpture", month="upcoming", user\_ratings="high")]} \\
\hline
CoALM-70B \textcolor{red}{\(\times\)}& \texttt{[find\_exhibition(location="New York", art\_form="modern sculpture", month="upcoming", user\_ratings="high")]} \\
\hline
\soce{} \textcolor{green}{\(\checkmark\)} & \texttt{[find\_exhibition(location='New York City, NY', art\_form='sculpture', user\_ratings='high')]} \\
\hline
\end{tabular}
\caption{Comparison of model outputs for task simple\_267 in BFCL v3.}
\label{tab:tasksimple}
\end{table*}

\section{Souping with Large Language Models with Mixture-of-experts architecture}
We conducted some baseline experiments on 6 models finetuned on Qwen/Qwen3-30B-A3B-Instruct-2507~\cite{qwen3_30b_a3b_instruct_2507} and 5 models Llama-4-Scout-17B-16E~\cite{llama4_scout_17b_16e} on the BFCL benchmark. We were able to find at least one souped candidate that outperformed all of the candidates. In case of checkpoints finetuned on Qwen, the category experts were not easy to distinguish, therefore, we observed only marginal improvements with the best individual candidate's average accuracy at 67.68$\%$ and the best souped model 68.48$\%$. It was easier to find distinct category experts with the checkpoints finetuned with Llama4 models and we saw that the principles of \soce{} could be applied. The best individual candidate's average accuracy at 58.51$\%$ and the best souped model 61.72$\%$. We only publish baseline results with pairwise souping as proof of concept that souping can be applied to large language models with dense as well as mixture-of-experts architectures, focusing on the former and leaving a more comprehensive study for the latter to future works.

\section{Licenses of Open Sourced Models} 
In this study, we have used several Open Sourced models. We have verified that these models are released under either Creative Commons Attribution-NonCommercial or Apache 2.0 licenses. 
\end{document}